\documentclass[pdflatex,sn-mathphys-num]{sn-jnl}

\usepackage{graphicx}%
\usepackage{subcaption}
\usepackage{multirow}%
\usepackage{amsmath,amssymb,amsfonts}%
\usepackage{amsthm}%
\usepackage{mathrsfs}%
\usepackage[title]{appendix}%
\usepackage{textcomp}%
\usepackage{manyfoot}%
\usepackage{booktabs}%
\usepackage{algorithm}%
\usepackage{algorithmicx}%
\usepackage{algpseudocode}%
\usepackage{listings}%
\usepackage{rotating}
\usepackage{comment}
\usepackage{rotating}
\usepackage{array}
\usepackage{bm}
\usepackage{makecell}

\usepackage{xcolor}

\newcommand{\fede}[1]{\textcolor{orange}{\textbf{#1}}}

\theoremstyle{thmstyleone}%
\theoremstyle{thmstyletwo}%

\theoremstyle{thmstylethree}%

\renewcommand{\arraystretch}{1.4}

\begin{document}

\title[Explainable cluster analysis]{Explainable cluster analysis: a bagging approach}


\author[1]{\fnm{Federico Maria} \sur{Quetti}}\email{federicomaria.quetti01@universitadipavia.it}
\equalcont{Both authors contributed equally to this work.}

\author*[2]{\fnm{Elena} \sur{Ballante}}\email{elena.ballante@unipv.it}
\equalcont{Both authors contributed equally to this work.}

\author[2]{\fnm{Silvia} \sur{Figini}}\email{silvia.figini@unipv.it}

\author[3]{\fnm{Paolo} \sur{Giudici}}\email{paolo.giudici@unipv.it}

\affil[1]{\orgdiv{Department of Mathematics}, \orgname{University of Pavia}, \orgaddress{\city{Pavia}, \postcode{27100},  \country{Italy}}}

\affil[2]{\orgdiv{Department of Social and Political Sciences}, \orgname{University of Pavia}, \orgaddress{\city{Pavia}, \postcode{27100},  \country{Italy}}}

\affil[3]{\orgdiv{Department of Economics and Management}, \orgname{University of Pavia}, \orgaddress{\city{Pavia}, \postcode{27100},  \country{Italy}}}


\abstract{A major limitation of clustering approaches is their lack of explainability: methods rarely provide insight into which features drive the grouping of similar observations.
To address this limitation, we propose an ensemble-based clustering framework that integrates bagging and feature dropout to generate feature importance scores, in analogy with feature importance mechanisms in supervised random forests. By leveraging multiple bootstrap resampling schemes and aggregating the resulting partitions, the method improves stability and robustness of the cluster definition, particularly in small-sample or noisy settings.
Feature importance is assessed through an information-theoretic approach: at each step, the mutual information between each feature and the estimated cluster labels is computed and weighted by a measure of clustering validity to emphasize well-formed partitions, before being aggregated into a final score.
The method outputs both a consensus partition and a corresponding measure of feature importance, enabling a unified interpretation of clustering structure and variable relevance. Its effectiveness is demonstrated on multiple simulated and real-world datasets.}

\keywords{Unsupervised learning,  Clustering, Bootstrapping, Bagging, Dunn Index, Mutual Information}



\maketitle

\section{Introduction}

A prevalent limitation of many machine learning models is their lack of explainability in terms of variable importance, as they  typically operate as a black box \cite{Lundberg2017}. Among machine learning methods, clustering is the process of grouping data points based on their similarity with respect to certain features, aiming to uncover inherent patterns or structures within the dataset. In the field of clustering many literature methods provide little insight into which data features drive the grouping of similar observations. 
Most existing research has focused on improving partition quality or tailoring clustering algorithms to specific structural assumptions, rather than providing criteria to assess the relevance of features. A first line of work refines benchmark partitional methods such as K-means \cite{jmac1967, lloyd1982least} to enhance within-cluster compactness and between-cluster separation \cite{ng2001spectral, dhillon2004kernel}. A second direction emphasizes robustness to noise and non-spherical cluster shapes, as in density-based approaches like DBSCAN \cite{dbscan}. Model-based clustering methods, including Gaussian mixture models, instead aim at coherence with an underlying probabilistic model \cite{mclachlan2000finite}. 
Complementing these approaches, recent work has explored ensemble clustering, in which the partitioning algorithm is repeatedly applied to resampled or perturbed versions of the data \cite{monti2003consensus, nguyen2007consensus, ghosh2011cluster}. This strategy improves the robustness and stability of the resulting partitions, mitigating sensitivity to noise, initialization, and algorithmic randomness.\\
Instead of directly assessing feature relevance, a common strategy is to address the related problem of feature selection: several methods incorporate embedded feature weighting or sparsity constraints directly into the clustering objective, thereby selecting variables during partition construction \cite{nino2021feature, oskouei2025feature}.
A growing body of literature has focused on variable selection in clustering: see
\cite{dy2004feature} and \cite{solorio2020review} for a review. 
Among these contributions, several works address feature extraction using information-theoretic methods \cite{gokcay2002information, torkkola2003feature, liu2009feature, faivishevsky2012unsupervised}. 
However, these approaches are primarily aimed at improving clustering performance by feature space adaptation, rather than providing a direct quantitative assessment of the individual contribution of each feature to the resulting partition. While variable selection in clustering has received considerable attention, comparatively fewer studies have focused on developing intrinsically interpretable clustering approaches \cite{interpretClust}, to the best of our knowledge. Existing proposals can be broadly categorized into several groups, following \cite{interpretClust}. A prominent line of research relies on rule-based clustering methods \cite{ruleClust}: examples are
the Unsupervised Random Forest (URF) \cite{URF} and the Clustering Using Binary Trees (CUBT) algorithm \cite{CUBT}, which are decision-tree–based and whose construction and hierarchical structure provide a natural form of interpretability. Other approaches include  optimization-based extensions of K-means that incorporate feature weighting mechanisms, for instance through entropy-based regularization \cite{jing2007entropy} or multilevel weighting schemes \cite{chen2011tw}. Lastly, we mention graph-based clustering methods \cite{belkin2001laplacian}, which capture complex geometric relationships in the data through latent space projections and allow for Laplacian-based scores for feature importance assessment. \\
In this paper, we introduce an explainable clustering framework grounded in information theory, leveraging mutual information, in line with previous contributions on both unsupervised and supervised methods \cite{franccois2007resampling, vergara2014review}.
This approach enables interpretable feature-importance scoring while offering a flexible, nonparametric methodology that can be applied to enhance any clustering algorithm.
Our framework takes a partitioning algorithm and enhances it in two ways: by estimating a mutual-information-based importance score for each feature, and by deriving an optimal consensus partition. Both tasks are carried out using a bootstrap aggregation (bagging, \cite{breiman1996bagging}) scheme that leverages multiple clustering realizations. One of the advantages of bagging lies in allowing to perform evaluations and operate independently on each bootstrap replica, in an ensemble framework. As introduced for the supervised approach in the Random Forest model \cite{breiman2001random}, a common method to decorrelate results and improve aggregation relies on feature dropout, where the operations are performed on random subspaces. In this work, a clustering approach is applied on each replica defined through subspace selection. A robust ensemble estimation of both cluster labels and of their associated mutual information with respect to each individual feature is achieved: the proposed importance score is evaluated based on aggregation of these information-theoretic values. Various resampling schemes are tested for comparison. 
As a baseline, we take replicas without data resampling, relying solely on initialization and algorithmic randomness for variability. We implement Efron’s bootstrap, treating the dataset as a realization of an underlying distribution, and use it to obtain a robust estimation of the unknown distribution \cite{efron1992bootstrap}.
Lastly, one of the proposed methods consists in extending the Bayesian Bagged Clustering (BBC) scheme proposed in \cite{quetti2025bayesian}.
The BBC algorithm stands on a Bayesian resampling framework, called proper Bayesian bootstrap (pBB) \cite{muliere1996bayesian}, which provides a mechanism for integrating prior knowledge into the clustering process. Such a  prior-informed scheme can be helpful to weigh features in a different manner, e.g. in consideration of their reproducibility. Aside from computing the importance scores for each feature, all resampling variants of the proposed method are used to output a consensus partition from the clusterings obtained for ensemble replications \cite{monti2003consensus, nguyen2007consensus, ghosh2011cluster}. 

The rest of the paper is structured as follows. In Section~\ref{sec:background}, the literature methods included in the proposal are detailed. The novel methodology is introduced in Section~\ref{sec:proposed_methodology}, and its results are described in Section~\ref{sec:results}. In Section~\ref{sec:conclusions}, conclusions and future directions are drawn.

\section{Background}\label{sec:background}
In this section, we review the key background elements of our proposed methodology.

\subsection{Clustering}
\label{clustering}
Clustering is the process of grouping data points based on their similarity with respect to certain features, aiming to uncover inherent patterns or structures within the dataset. Methods existing in the literature \cite{jain1999data} differ on the approach chosen and with respect to the similarity definition used. The most common ones are: partitional methods, aimed at dividing data as to minimize a cost function iteratively \cite{Elements2009}; hierarchical methods, which organize data into a tree-like nested structure, according to a similarity measure \cite{hierarchical}; density methods, assigning groups by separating different density regions \cite{dbscan}; model-based methods, which optimize the fit with an underlying probabilistic model \cite{mclachlan2000finite}.\\
In this work, grounded in the BBC method, the clustering algorithm of choice is the K-means algorithm \cite{jmac1967} with Euclidean distance, as a paradigmatic partition-based method. 
K-means is an iterative algorithm which, at each iteration, groups $n$ observations into $K$ clusters, where $K$ is the main hyperparameter of the method. Once fixed, the algorithm searches an optimal clustering over the space of partitions $\{1,\dots,K\}^n$ by minimizing the within-cluster sum of squares (WSS):
\begin{equation*}
\label{wssKm}
    WSS = \sum_{k=1}^{K} \sum_{i \in C_k} \left\| x_i - \mu_k \right\|^2,
\end{equation*}
where $C_k$ is the set of points in cluster $k$ and $\boldsymbol{\mu}_k$ their centroid. The centroids are randomly chosen in the first iteration and updated in further iterations as the mean of the observations in each cluster $C_k$. \\



\subsection{Bootstrap aggregation: bagging}
\label{bagging}

Bootstrap is a statistical resampling technique used to estimate the distribution of a statistic, providing an approximation to its empirical distribution given a set of data points. Formally, given $\{X_1,\dots,X_n\}$ i.i.d. random variables with same distribution as $X$, we are interested in estimating the distribution of a functional $\Phi(F,X)$, depending on the cumulative distribution function (cdf) $F$ of $X$. \\
A first method to get an estimate of $F$ was proposed by Efron \cite{efron1992bootstrap}, obtained by generating replications with replacement from the sample. This procedure is equivalent to drawing a weights vector for observations sampled from a Multinomial distribution The estimate is then defined as:
\begin{equation*}
    F^*(x)= \sum_{i=1}^n \frac{w_i}{n}  \mathbb{I}[X_i \leq x],
\end{equation*}
with $(w_1, \dots, w_n) \sim \textit{Mult}(n, \frac{1}{n}\mathbf{1}_n)$ \ and \  $\mathbb{I}[X_i \leq x]$ the indicator function. \\
An alternative bootstrap method was proposed by Rubin in \cite{rubin1981bayesian}, and named the Bayesian bootstrap: the method is similar to Efron's but differs in the definition of the weights, drawn from a Dirichlet distribution. The approximation of the cdf is:
\begin{equation*}
F^*(x) = \sum_{i=1}^{n} w_i \mathbb{I}[X_i \leq x],
\end{equation*}
where $(w_1, \dots, w_n) \sim D(\mathbf{1}_n)$. 
The two bootstrap methods are asymptotically equivalent \cite{Lo1987}, and first order equivalent from the predictive point of view, as the conditional probability of a new observation is estimated by only using observed values from the sample \cite{galvani2021}. 

One of the resampling schemes employed in this paper, at the core of the BBC method, is the proper Bayesian bootstrap (pBB) \cite{muliere1996bayesian}, which generalizes the previously described methods by integrating prior knowledge into the resampling process. 
The assumption of the method is the true distribution itself be generated by a Dirichlet process $\mathcal{DP}$ \cite{ferguson1973bayesian}. 
Prior knowledge is introduced by defining a prior Dirichlet process, with baseline distribution $F_0$, and with a certain amount of confidence, $k$.
Given a random sample from data $\{x_1, \dots, x_n\}$ with its empirical distribution $F_n$, and the prior distribution generated by the Dirichlet process $F \sim \mathcal{DP}(kF_0)$, the resulting posterior distribution function is generated from the conjugate process $\mathcal{DP}((k + n)G_n)$,
where the baseline distribution function is given by:
\[
G_n = \frac{k}{k + n}F_0 + \frac{n}{k + n}F_n.
\]
Bootstrap datasets are built by sampling from the posterior distribution generated by the posterior process. Formally, 
$X_1, \dots, X_m \mid G \stackrel{iid}{\sim} G$, where $G \sim \mathcal{DP}((k + n)G_n)$. \\

Directly defined from the aggregation of bootstrap copies, bagging is an ensemble learning method introduced by Breiman \cite{breiman1996bagging} in a supervised learning setting to mitigate the variance of unstable models and improve prediction accuracy by combining multiple models. The procedure relies on generating $B$ bootstrap samples from the original training dataset and fitting a separate model on each resampled dataset.  By aggregating these predictions, bagging stabilizes the estimation and improves predictive accuracy, particularly for high-variance methods such as decision trees. \\
The bagged prediction in regression settings is defined as the average of the bootstrap predictions:
\begin{equation*}
\hat f_{\mathrm{Bag}}(x)\ =\ \frac{1}{B}\sum_{b=1}^B \hat f^{*(b)}(x),
\end{equation*}
where $B$ denotes the number of bootstrap replicas and $\hat f^{*(b)}(x)$ is the prediction obtained on the $b$-th bootstrap replica. In classification settings, the aggregation is usually carried out by majority voting.

\subsection{Bagged clustering}
\label{sec:bagged_clustering}
Several approaches involving bagging are present in the field of clustering. One of the first examples was introduced in the work of Dudoit and Fridlyand \cite{dudoit2003bagging}, inspired by the idea of Leisch \cite{Leisch}: a bagging approach is used to perform clustering through both partition-based and hierarchical algorithms (after selection of the number of clusters parameter), in order to improve the stability of the results.
In \cite{dudoit2003bagging}, they define the algorithm BagClust1, which assigns through bagging an aggregated cluster label to each observation based on the majority vote, while also retrieving a fuzzy-type result by recording the cluster memberships. \\
In a recent proposal \cite{quetti2025bayesian}, the bagged clustering algorithm BagClust1 of \cite{dudoit2003bagging} is adapted to the proper Bayesian bootstrap scheme, leading to Bayesian Bootstrap Clustering (BBC). The BBC algorithm \cite{quetti2025bayesian} first performs a partition-based clustering to elicit the prior for the proper Bayesian bootstrap. Subsequently, a bagging scheme is applied, in which the resampling step is carried out through pBB, and K-means is fitted to each bootstrap replica. 
Finally, the clustering memberships obtained across replicates are aggregated to produce cluster allocation probabilities; these, in turn, produce hard assignments obtained from majority voting. \\
Out of the several aggregation methodologies that have been introduced in the literature \cite{nguyen2007consensus, ghosh2011cluster}, henceforth in this work we will focus on consensus via optimization of a loss function (see Section~\ref{sec:vi}). This approach circumvents the computational hurdle of relabeling the clusters between steps of the ensembling procedure, which affected greatly the aforementioned algorithms, and allows for a clear theoretical interpretation (see Section~\ref{theoretical_justification}).

\subsection{Validation}
\label{validation}
Several measures of the quality of hard clustering methods exist, divided into two categories:  internal indexes, which evaluate the quality of the clustering based on the computed labels, and external indexes, which compare the computed labels with a ground truth \cite{validation}.\\
In our proposal, we employ, as an internal index,  the Dunn index:
$$\mathrm{Dunn}(x_1, \dots, x_n, \ell) = \frac{\displaystyle \min_{1 \le i < j \le K} d(C_i, C_j)}
       {\displaystyle \max_{1 \le k \le K} \Delta(C_k)},$$
where $x_i$ are the $p$ dimensional data points, and $\ell$ the $n$-dimensional labels vector, $C_j$ the $j$-th cluster,  $d(C_i, C_j) = \min_{x \in C_i, y \in C_j} \|x - y\|$ as an inter-cluster distance, and $\Delta(C_k) = \max_{x, y \in C_k} \|x - y\|$ as an intra-cluster distance. The index takes values in $[0,+\infty)$, where a value of 0 corresponds to the worst case of absence of cluster separation, while larger values indicate increasingly well-defined clusters. 

Concerning external validation indexes, a possible choice is the Rand index \cite{rand1971objective}: 
 $$\mathrm{Rand}(\ell, \ell_{true}) = \frac{a + b}{\binom{n}{2}},$$
 where $a$ denotes the number of pairs of elements in the same cluster, and $b$ the number of pairs in different clusters, in the estimated ($\ell$) and ground-truth ($\ell_{true}$) partition. 
 Since the Rand index does not correct for agreement due to chance, we consider its adjusted version, the Adjusted Rand Index (ARI) \cite{hubert1985comparing}, which ranges between -1 and 1, has expected value 0 for random partitions and equals 1 for identical clusterings.
As a further validation measure, we consider the Fowlkes–Mallows index (FMI) \cite{fowlkes1983method}. Let $TP$ denote the number of pairs assigned to the same cluster in both the predicted and the true partition, $FP$ the number of pairs predicted in the same cluster but belonging to different ground truth clusters, and $FN$ the number of pairs in the same ground truth cluster but not predicted as such. The FMI is then defined as
\[
\mathrm{FMI}(\ell,\ell_{true}) = \sqrt{ \frac{TP}{TP + FP} \cdot \frac{TP}{TP + FN} }.
\]


\subsection{Mutual information}
\label{mutual_information}

To quantify the importance of features in producing a partition, we propose to employ the Mutual Information (MI), based on the notion of entropy for random variables \cite{shannon1948mathematical, cover1999elements}. 
Given a pair of discrete random variables $(X,Y)$, whose joint probability distribution is $p_{X,Y}(x,y)$ and marginals $p_X(x),\ p_Y(y)$, their mutual information is:
\begin{equation}
I(X;Y) = H(X)-H(X|Y) =H(X)-\sum_{\mathrm{supp}(Y)}p(y)H(X|Y=y).
\label{mutual}
\end{equation}
where \[H(X)= -\sum_{\mathrm{supp}(X)}p(X)\log_2p(X)\] is the Shannon entropy. 
The properties of the mutual information, following from its definition in Equation~\ref{mutual}, carry a clear interpretation in the clustering context. Let $Y$ be associated to a cluster labeling vector. Feature $X_i$ is considered more informative, and thus more important for clustering, than feature $X_j$ if $I(X_i;Y)>I(X_j;Y)$. For a feature variable $X$ it holds that:

\begin{itemize}
    \item $I(X;Y) \geq 0$: non-negativity implies that information about the labels cannot increase the uncertainty about the feature, and vice versa.
    
    \item $I(X;Y) = I(Y;X)$: the information that $Y$ provides about $X$ is equal to the \emph{information that $X$ provides about $Y$}. This symmetry justifies estimating mutual information via $I(X;Y) = H(X) - H(X \mid Y)$, which is computationally convenient in our setting, as it relies on estimating the marginal distribution $p_X(x)$ and the conditional distributions $p_X(x \mid Y=\ell)$ for each cluster.
\end{itemize}
For continuous variables in finite datasets, the computation of these quantities is necessarily approximate, as it involves continuous distributions. Kernel Density Estimation (KDE) \cite{parzen1962estimation, kozachenko1987sample, davis2011remarks} can be employed to empirically estimate\footnote{Histogram- and kernel-based discretizations require the specification of a tuning parameter; in our KDE implementation, the bandwidth is chosen using Silverman’s rule of thumb.} 
the probability distribution of a vector, and, thus, the entropy and the mutual information, thereby allowing the implementation of Equation~\ref{mutual}.


\subsection{Variation of information}
\label{sec:vi}

In order to define a consensus partition 
from a collection of different clusterings, a criterion is required to select an optimal partition among the candidates. This selection, in an optimization framework, can be formulated by adopting the Variation of Information (VI) \cite{meilua2003comparing, wade2018bayesian} in the definition of a loss function. The VI is a metric on the space of partitions of a fixed dataset and is defined as:
\begin{equation}
\label{eq:VI}
    VI(\ell_1, \ell_2) = H(\ell_1) + H(\ell_2) - 2I(\ell_1, \ell_2),
\end{equation}    
where $\ell_1,\ell_2$ are elements of the partition space of the same dataset $\mathbf{X}$.
In principle, the optimal consensus partition is obtained by solving
$\ell^* = 
\mathrm{argmin}_{\ell_2 \in \mathcal{L}} \mathbb{E}\left[ VI(\ell_1, \ell_2) \mid \mathbf{X}\right],$ which amounts to evaluating the Fréchet mean in the space of all possible partitions.
To mitigate the computational cost of this optimization, we restrict the search space to the set of observed partitions, effectively selecting the restricted Fréchet mean in the sampled partition space, with respect to the VI metric in Equation~\ref{eq:VI}. \\
Finally, we note that the variation of information can be computed from the contingency (co-occurrence) matrix of data points. As a consequence, it is invariant under relabeling of clusters and can be efficiently computed in downstream analyses.

\section{Proposed methodology}
\label{sec:proposed_methodology}
In this section we first detail the steps of our proposed methodology, and then give a formal justification of the design choices.

\subsection{Proposal}
\label{proposal}

Our proposal combines the previously described techniques, and is detailed by the following steps.

\begin{enumerate}
\setcounter{enumi}{-1} 
    \item \textbf{Preprocessing}. Standardization of each feature, to avoid scale dependence within the clustering procedure. This is due to the choice of partitioning algorithm and distance function (K-means with euclidean distance). 
    \item \textbf{Bootstrap}. The first part of the proposed method consists in a bootstrap resampling scheme. Three methods, of increasing complexity, are compared, namely: cloning of the dataset (no resampling); Efron bootstrap resampling; proper Bayesian bootstrap resampling, as described in Section~\ref{sec:bagged_clustering}.
    
    \item \textbf{Clustering on random subspaces (feature dropout)}. K-means is performed on each resampled replica, after random subspace selection: similarly to random forest \cite{breiman2001random}, $m < p$ features are randomly selected and used for partitioning in each round. As replicas are obtained randomly, clustering typically involves a subset of the original dataset. To recover assignments to the full set of original points, we adopt a nearest centroid assignment rule: each point is assigned to the cluster with closest centroid.

    \item \textbf{Validation step}. The Dunn index, defined in Section~\ref{validation}, is employed as an internal cluster quality index for the original dataset and its relative cluster labels, obtained at step 2. The index is then used as a weight for the final computation of the feature importance score and for aggregation of the clustering results.

    \item \textbf{Mutual information computation}. 
    The cluster label assignments 
    are employed to compute the mutual information with each individual feature involved in step 2, as described in Section~\ref{mutual_information}, within the original dataset. For clarity of results, we normalize the value to 1 over all features,
    bounding it in the interval $[0,1]$. 
    
    \item \textbf{Aggregation of results}. 
    Feature importance scoring is computed as the weighted average of the importance values from step 4, only over the bootstrap replicas in which the feature is involved, using the Dunn indexes obtained at step 3 as weights. 
    The consensus clustering is obtained by evaluating the Fréchet mean, weighted by the Dunn index, in the space of computed partitions (see Section~\ref{sec:vi}). 
    
\end{enumerate}

\subsection{Theoretical interpretation}
\label{theoretical_justification}
Consensus partition evaluation and feature scoring are treated as distinct decision problems. The consensus partition provides a representative clustering out of the realized observations, while feature relevance is assessed independently by aggregating importance scores derived from mutual information estimates across the ensemble. In the treatment of both problems, internal clustering quality weighting acts as an importance-sampling mechanism over the space of partitions, emphasizing reliable cluster structures while down-weighting the poorly separated ones.
Let:
\begin{itemize}
  \item $\mathbf{X} \in \mathbb{R}^{n \times p}$ denote the data matrix, with features $X_1, \dots, X_p$;
  \item $b = 1, \dots, B$ be the clustering runs index;
  \item $S_b \subseteq \{1,\dots,p\},\ |S_b| =  m$, be the subspace of feature indexes randomly selected in the $b$th clustering run;
  \item $\mathcal{B}_j = \{ b \in \{1,\dots,B\}: j \in S_b \}$ denote the set of runs in which feature $X_j$ is selected;
  \item $\ell^{(b)} \in \{1,\dots,K\}^n $ denote the cluster labels obtained in run $b$ (e.g.\ via K-means on a random subspace), element of the partitions set $ \mathcal{L}$;
  \item $w_b \ge 0$ be an internal quality index for the dataset and its partition on run $b$ (e.g. Dunn index of the full dataset $\mathbf{X}$ with labels $\ell^{(b)}$).
\end{itemize}
\textbf{Consensus partition.} 
The stochastic nature of the proposed procedure induces a probability distribution over partitions of the $n$ observations. Each clustering run $b$ yields a partition $\ell^{(b)}$, and the collection $\{\ell^{(b)}\}_{b=1}^B$ can be regarded as a finite sample from an implicit distribution over $\mathcal{L}$, determined by the combined effects of data resampling (in the Efron and BBC variants at step 1), feature-subspace sampling (in step 2), and algorithmic randomness (intrinsic to K-means at step 2). Individual realizations represent random draws from an ensemble: 
the task of finding the best partition from such a distribution can thus be framed as seeking a consensus from the realized partitions set \cite{wade2018bayesian, jain2015asymptotic}.
We adopt a decision-theoretic approach and define the consensus partition as the one minimizing an expected loss over elements of $\mathcal{L}$. Let $d(\cdot,\cdot)$ denote a distance between partitions; in line with our information-theoretic framework, we choose the variation of information (VI) \cite{meilua2003comparing}, described in Section~\ref{sec:vi}, as distance in the loss function definition. The consensus partition $\ell^\star$ is then defined as:
\begin{equation}
\label{eq:consensus_partition}
\ell^\star
=
\operatorname*{argmin}_{\ell \in \{\ell^{(b)}\}}
\sum_{b=1}^B w_b \, d_{\mathrm{VI}}\!\left(\ell, \ell^{(b)}\right).
\end{equation}
Equation~\ref{eq:consensus_partition} defines a Fréchet mean of the partitions under the VI distance, weighted by the internal quality of the clustering obtained in run $b$.
This definition selects a clustering that is, on average, closest to the ensemble of sampled clusterings under the chosen loss, with higher-quality realizations contributing more strongly to the decision. 
As such, it captures the clustering structure that is most consistently supported across high-quality runs, while attenuating the influence of poor or even degenerate outcomes, which are expected due to random dropout of relevant features. 
\\

\textbf{Feature importance scoring.} 
Consider now the proposed feature scoring. Formally, at run $b$ a subset of $m$ feature indexes $S_b \subset \{1,\dots,p\}$ is sampled, yielding a restricted data matrix $\mathbf{X}_{S_b} \in \mathbb{R}^{n \times m}$. This is resampled and clustered, leading to its corresponding partition $\ell^{(b)}$. 
For a given feature $X_j \subset \mathbf{X}_{S_b}$, the mutual information with the obtained cluster labels is:
\[
I(X_j; \ell^{(b)} \mid S_b).
\]
We remark that this quantity is retained only if $j \in S_b$: in runs for which $j \notin S_b$ it is not used for computation of feature importance for $X_j$, since the associated partitions do not encode information attributable to that feature. We further normalize this value to the range [0,1], and denote it as $\tilde{I}(X_j; \ell^{(b)} \mid S_b)$.
The final feature importance score is defined as the aggregated value of $\tilde{I}$ across these realizations, weighted by the clustering quality $w_b$:

\begin{equation}
\label{eq:feature_importance}
\mathrm{FI}_j
=
\frac{\sum_{b \in \mathcal{B}_j}
w_b \tilde{I}(X_j; \ell^{(b)} \mid S_b)}{\sum_{b \in \mathcal{B}_j} w_b}.
\end{equation}
The estimator in Equation~\ref{eq:feature_importance}
can be interpreted as a quality-weighted ensemble average of individual scores $\tilde{I}(X_j;\ell^{(b)} \mid S_b)$, computed under heterogeneous probability laws. Each term is evaluated with respect to the empirical joint distribution induced by the clustering mechanism associated with run $b$, which is itself determined by the randomly selected feature subspace $S_b$, the resampling procedure, and the initialization of the clustering algorithm. 
The aggregation is further restricted to the subset of runs \(\mathcal{B}_j\) in which feature \(X_j\) is included in the construction of the partition. Normalizing by \(\sum_{b \in \mathcal{B}_j} w_b\) therefore defines an effective sampling measure over clustering realizations, conditional on the active participation of feature \(X_j\). 

\section{Results}\label{sec:results}

To demonstrate the effectiveness and robustness of our proposal, as well as comparing it with previous studies, we consider simulated and real datasets. The correct number of clusters in each case is assumed to be known. \\
Two sets of results are reported, illustrating the performances of the algorithm as a clustering method and in terms of explainability. All the results are shown with different choices of the main hyperparameter of our method, called \textit{subspace size} or \textit{dropout parameter} and indicated as $m$, in order to discuss the robustness of the method.\\
\textbf{Clustering performances.} 
The clustering quality results for each method are evaluated in comparison to the ground truth in terms of adjusted Rand (ARI) and Fowlkes-Mallows (FM) index (see Section~\ref{validation}).\\ 
As benchmark clustering method, we employ the most common algorithms in applied contexts:  K-means \cite{jmac1967} and Hierarchical clustering \cite{hierarchical}.\\
\textbf{Explainability.} The advantages of the proposed method are shown in comparison with the existing cluster explainability methods in the literature.
The compared methods are selected from the recent review \cite{importanceCUBT}, and are:
\begin{itemize}
 \item \textbf{URF}$_{\text{\textbf{Gini}}}$: Unsupervised Random Forest, where variable importance is derived from the Gini index associated with tree splits on each individual variable, measured as the average Gini decrease across all trees \cite{URF}. \\
\item \textbf{URF}$_{\text{\textbf{perm}}}$: Unsupervised Random Forest, where variable importance is quantified as the difference between the predictive error and the error obtained after randomly permuting each individual variable, averaged across all trees \cite{URF}. \\
 \item \textbf{Clustering Using Binary Trees (CUBT)}:
 evaluates feature importance by assessing, under a surrogate tree-growing criterion, how much each variable contributes to reconstructing the original cluster partition through the induced tree structure, see \cite{importanceCUBT}.\\ 
 \item \textbf{TWo level KMeans (TWKM)}: variable importance is described by the feature weights in the optimization procedure of the algorithm \cite{chen2011tw}. 
 Extending standard K-means, the method generalizes the loss function to account for feature and group weights, as well as adding two entropy terms (one for single features and one for groups). 
 \item \textbf{Laplacian Score (LS)}: variable importance is measured using the Laplacian Score \cite{belkin2001laplacian}, which is computed based on the proximities between observations in a $k$-nearest neighbor graph. The method assumes that two observations are related if they are connected in such graph.

\end{itemize}

For each method used, greater values correspond to more important features, except for the Laplacian Score (LS) which is to be interpreted in the opposite way. Note that some of the methods considered in \cite{importanceCUBT} require a careful parameter tuning. This can be considered as a disadvantage, as it reduces reproducibility; instead, our proposals do not suffer from this limitation, by being robust on the parameter $m$.
For the number of replica parameters, we set as default $B=100$ replicas, as this choice provides a reasonable balance between robustness and computational cost. \\
In order to understand the role of resampling techniques in our study, the results are calculated using three different variants of the model: 
\begin{itemize}
\item \textbf{Basic}: no resampling method is applied, and all the data variability comes by the dropout mechanism.\\
\item \textbf{Efron}: the resampling method is the classical Efron's bootstrap. \\
\item \textbf{BBC}: the final proposal relies on the proper Bayesian bootstrap \cite{muliere1996bayesian}, as originally proposed in the BBC method \cite{quetti2025bayesian}. This method needs the specification of the prior weight parameter $\omega$, here chosen as $\omega = 0.1$ henceforth and validated by the sensitivity analysis in \cite{quetti2025bayesian}.
\end{itemize}
To facilitate the interpretation, all importance scores of literature methods are normalized between features, forcing their range in [0,1].

\subsection{Synthetic datasets}

\subsubsection{Illustrative example}
\label{sec:illustrative}

\begin{figure}[h]
    \centering    
    \includegraphics[width=\linewidth]{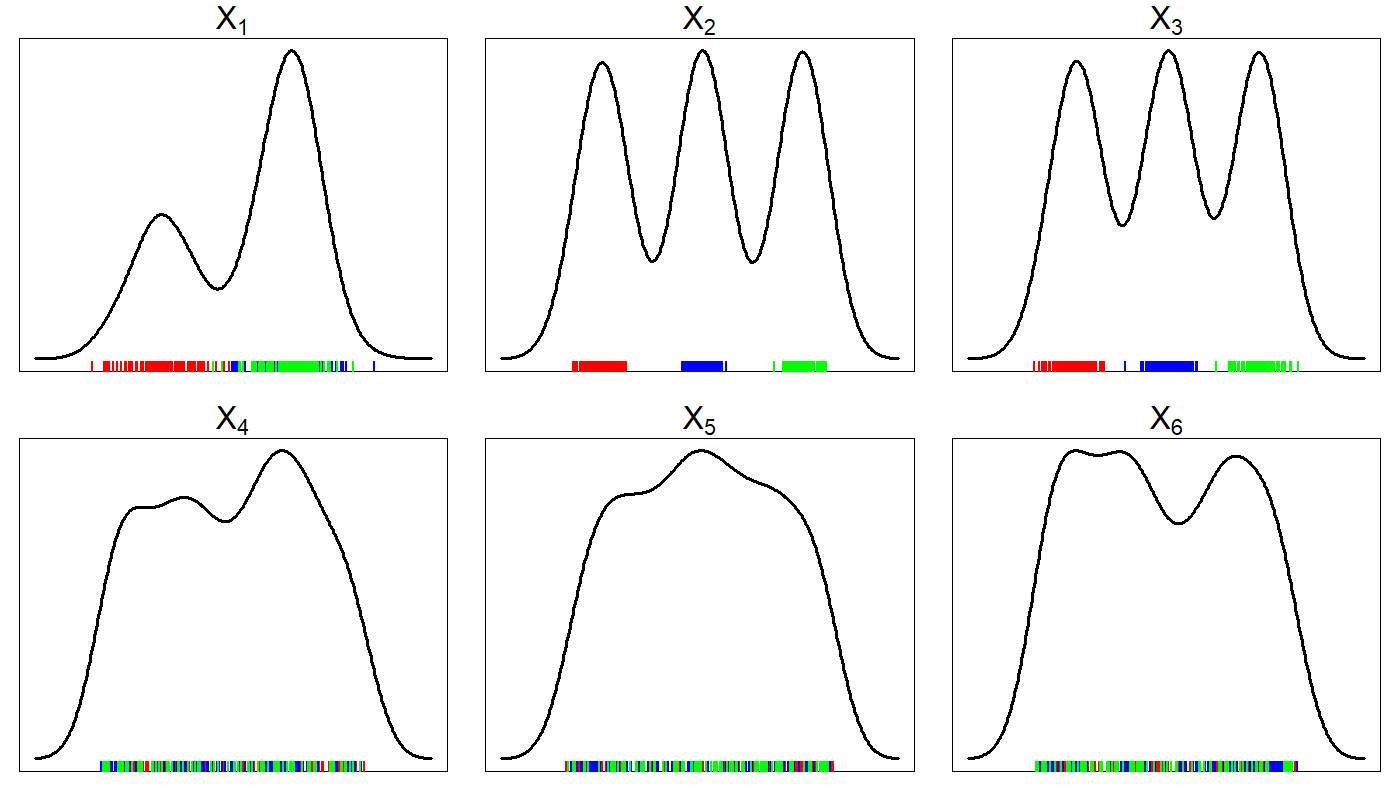}
    \caption{Kernel density estimation of each feature for a simulation of the synthetic dataset. A rugplot of the occurrences colored by true cluster label is shown.}
    \label{kde_toy}
\end{figure}

As an illustrative example, we consider a synthetic dataset with $p=6$ features. Two features are trimodal, one is bimodal, and the three remaining are uniform noise with different ranges. The true number of clusters is $K=3$, each one containing 99 observations. The data generation process is described in detail in Table~\ref{tab:feature_generation_ds1} and the distribution of data estimated via KDE is visually represented in Figure~\ref{kde_toy}.
\begin{table}[ht!]
\centering
\scalebox{1}{
\begin{tabular}{|p{1.25cm}|p{3.4cm}|p{7cm}|}
\hline
\textbf{Variable} & \makecell{\textbf{Type}/\textbf{Distribution}} & \textbf{Generation} \\
\hline
$X_1$ & \makecell{Gaussian \\ (cluster-dependent)} & First coordinate of a bivariate Gaussian $\mathcal{N}(\mu_i, \Sigma)$, where $\mu_i \in \{(0,0), (3,3), (3,6)\}$ is conditional on the cluster and  $\Sigma = \begin{pmatrix}0.5 & 0\\ 0 & 0.1\end{pmatrix}$, constant for all clusters. \\
\hline
$X_2$ & \makecell{Gaussian \\ (cluster-dependent)}  & Second coordinate  of the same bivariate Gaussian distribution as $X_1$. Note that, while two  cluster means are the same in the first coordinate, they are not in the second. For this reason, $X_1$ is bimodal, and $X_2$ trimodal.  \\
\hline
$X_3$ & \makecell{Gaussian \\ (cluster-dependent)}  & Sampled from three independent and well separated univariate Gaussian distributions, conditionally on the cluster:  $\mathcal{N}(1, 0.09)$, $\mathcal{N}(3, 0.09)$, $\mathcal{N}(5, 0.09)$ \\
\hline
$X_4$ & \makecell{Uniform in the range $[a, b]$} & Each cluster is sampled independently from a uniform distribution in $[a, b]$ where $a = \min(X_1, X_2)$, $b = \max(X_1, X_2)$. \\
\hline
$X_5$ & \makecell{Uniform in the range \\ $[\min X_1$, $ \max X_1]$} & Each cluster is sampled independently from a uniform distribution over the $X_1$ observed range. \\
\hline
$X_6$ & \makecell{Uniform in the range \\ $[\min X_2$, $ \max X_2]$}  & Each cluster is sampled independently from a uniform distribution over the $X_2$ observed range. \\
\hline
\end{tabular}}
\caption{Description of the feature generation process of the illustrative dataset. }
\label{tab:feature_generation_ds1}
\end{table}
The results in terms of clustering performance for different choices of the subspace size $m$ are reported in  Table~\ref{tab:clustering_synthetic_m}.  
The basic version of the method achieves comparable results, while the introduction of bootstrap methods (Efron's and pBB) surpasses the benchmark given by K-means and hierarchical clustering methods when dropout is present. This toy scenario shows how removing redundant features and introducing the weighting mechanism stabilizes performance.

\begin{figure}[ht!]
    \centering    
    \includegraphics[width=\linewidth]{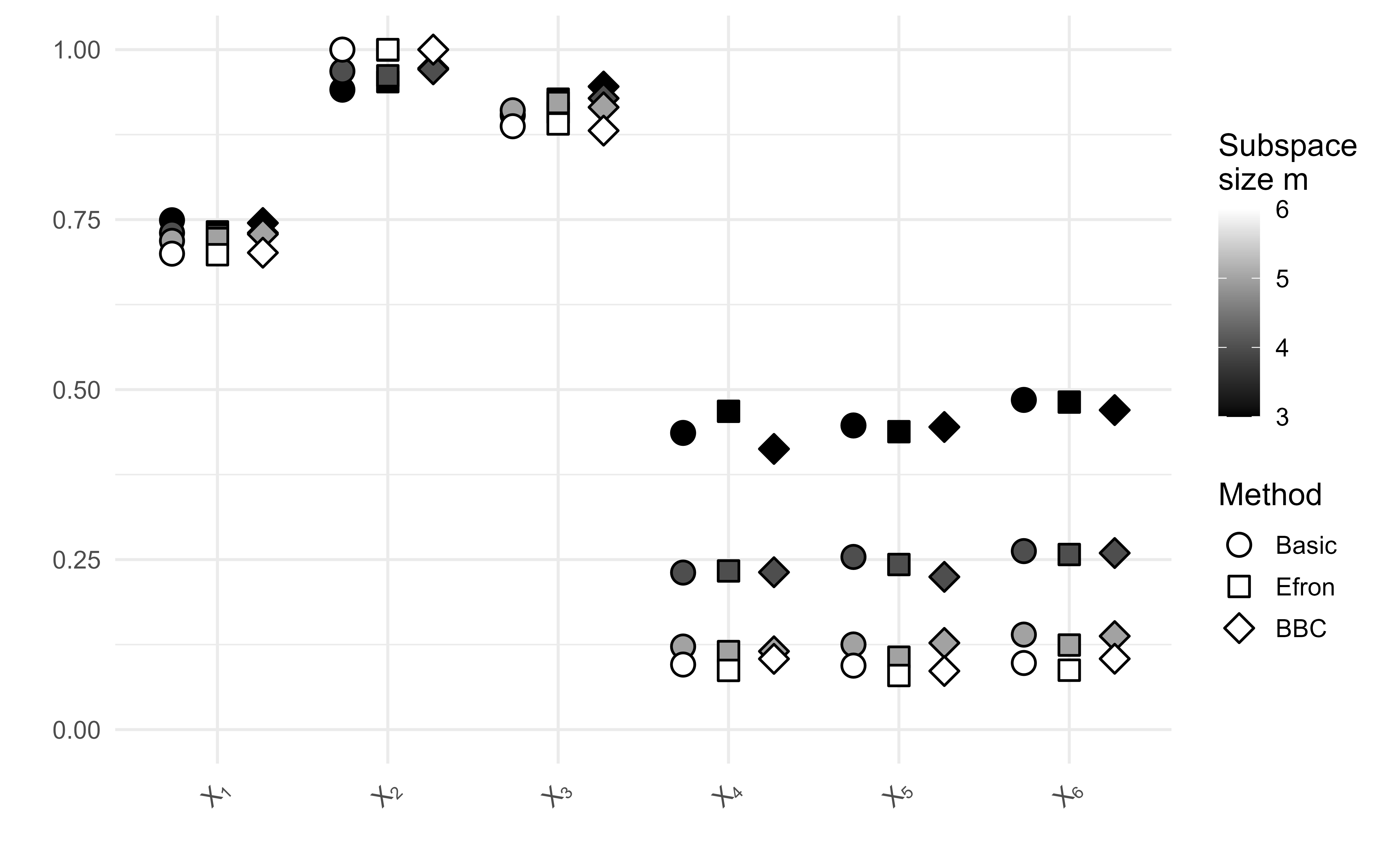}
    \caption{Comparison of the mean importance of each feature evaluated by our three proposals for different choices of subspace size, over 20 generations of dataset.}
    \label{fig:MeanVariableImportance_synthetic_per_m}
\end{figure}
 Considering the performances in terms of explainability, in Figure~\ref{fig:MeanVariableImportance_synthetic_per_m} we illustrate the comparison of the proposed feature importance scores for various choices of subspace size $m$.
 The results illustrate a stable ranking of features for decreasing $m$, where noisy features are impacted the most by the choice of the parameter, slightly reducing the distance with the informative features. This is due to the relatively higher importance assumed by features when the subspace size decreases, especially in such a low dimensional example. 
Given the robustness of the results concerning feature importance, in Table~\ref{tab:importance_BBC} the results obtained by the method by choosing $m=5$ are reported, and Table~\ref{tab:importance_lett} offers a comparison with literature methods. 


\begin{table}[t]
\centering
\begin{subtable}{\linewidth}
\centering
\begin{tabular}{cccccc}
\hline
$m$ & \textbf{K-means} & \textbf{Hierarchical} & \textbf{Basic} & \textbf{Efron} & \textbf{BBC} \\
\hline
3 & $0.55 \pm 0.16$ & $0.51 \pm 0.02$ & $0.76 \pm 0.28$ & $0.94 \pm 0.17$ & $1.00 \pm 0.00$ \\
4 & $0.55 \pm 0.16$ & $0.51 \pm 0.02$ & $0.49 \pm 0.12$ & $0.58 \pm 0.25$ & $0.88 \pm 0.22$ \\
5 & $0.55 \pm 0.16$ & $0.51 \pm 0.02$ & $0.51 \pm 0.12$ & $0.59 \pm 0.24$ & $0.50 \pm 0.03$ \\
6 & $0.55 \pm 0.16$ & $0.51 \pm 0.02$ & $0.50 \pm 0.04$ & $0.46 \pm 0.03$ & $0.51 \pm 0.03$ \\
\hline
\end{tabular}
\caption{Adjusted Rand index of the illustrative dataset.}

\label{tab:clustering_synthetic_m_adjRand}
\end{subtable}
\begin{subtable}{\linewidth
}
\centering
\begin{tabular}{cccccc}
\hline
$m$ & \textbf{K-means} & \textbf{Hierarchical} & \textbf{Basic} & \textbf{Efron} & \textbf{BBC} \\
\hline
3 & $0.71 \pm 0.10$ & $0.68 \pm 0.02$ & $0.85 \pm 0.17$ & $0.97 \pm 0.10$ & $1.00 \pm 0.00$ \\
4 & $0.71 \pm 0.10$ & $0.68 \pm 0.02$ & $0.69 \pm 0.07$ & $0.76 \pm 0.14$ & $0.92 \pm 0.14$ \\
5 & $0.71 \pm 0.10$ & $0.68 \pm 0.02$ & $0.69 \pm 0.07$ & $0.76 \pm 0.14$ & $0.67 \pm 0.02$ \\
6 & $0.71 \pm 0.10$ & $0.68 \pm 0.02$ & $0.68 \pm 0.01$ & $0.68 \pm 0.01$ & $0.68 \pm 0.02$ \\
\hline
\end{tabular}
\caption{Fowlkes-Mallows index of the illustrative dataset.}

\label{tab:clustering_synthetic_m_adjRand}
\end{subtable}
\caption{Clustering performance in terms of Adjusted Rand and Fowlkes-Mallows index of the illustrative dataset. Results are reported as mean $\pm$ standar deviation over 20 replications for different values of the subspace size $m$.}
\label{tab:clustering_synthetic_m}
\end{table}


\begin{table}[ht]
\centering
\begin{subtable}{\linewidth}
\centering
\begin{tabular}{lcccccc}
\hline
\textbf{Method} & $X_1$ & $X_2$ & $X_3$ & $X_4$ & $X_5$ & $X_6$ \\
\hline
Basic & 0.71$\pm$0.03 & 1$\pm$0 & 0.91$\pm$0.02 & 0.13$\pm$0.04 & 0.13$\pm$0.04 & 0.13$\pm$0.05 \\
Efron & 0.71$\pm$0.03 & 1$\pm$0 & 0.92$\pm$0.02 & 0.12$\pm$0.03 & 0.11$\pm$0.03 & 0.12$\pm$0.03 \\
BBC & 0.72$\pm$0.02 & 1$\pm$0 & 0.92$\pm$0.02 & 0.13$\pm$0.05 & 0.13$\pm$0.05 & 0.12$\pm$0.05 \\
\hline
\end{tabular}
\caption{Feature importance scores of the features for each resampling scheme. The subspace size is set as $m=5$.}\label{tab:importance_BBC}
\end{subtable}


\begin{subtable}{\linewidth}
\centering
\begin{tabular}{lcccccc}
\hline
\textbf{Method} & $X_1$ & $X_2$ & $X_3$ & $X_4$ & $X_5$ & $X_6$ \\
\hline
URF$_{\text{Gini}}$ &
$0.69\!\pm\!0.08$ & $0.96\!\pm\!0.06$ & $0.97\!\pm\!0.05$ &
$0.02\!\pm\!0.03$ & $0.02\!\pm\!0.02$ & $0.03\!\pm\!0.03$ \\

URF$_{\text{perm}}$ &
$0.78\!\pm\!0.06$ & $0.97\!\pm\!0.04$ & $0.97\!\pm\!0.05$ &
$0.02\!\pm\!0.03$ & $0.01\!\pm\!0.02$ & $0.02\!\pm\!0.02$ \\

CUBT &
$0.82\!\pm\!0.05$ & $0.98\!\pm\!0.03$ & $0.99\!\pm\!0.02$ &
$0.46\!\pm\!0.08$ & $0.47\!\pm\!0.09$ & $0.42\!\pm\!0.07$ \\

TWKM &
$0.72\!\pm\!0.45$ & $0.70\!\pm\!0.31$ & $0.19\!\pm\!0.13$ &
$0.18\!\pm\!0.38$ & $0.18\!\pm\!0.38$ & $0.26\!\pm\!0.44$ \\

LS &
$0.78\!\pm\!0.01$ & $0.72\!\pm\!0.01$ & $0.72\!\pm\!0.01$ &
$1.00\!\pm\!0.00$ & $1.00\!\pm\!0.00$ & $1.00\!\pm\!0.00$ \\
\hline
\end{tabular}
\caption{Feature importance scores for literature methods.}\label{tab:importance_lett}
\end{subtable}
\caption{Comparison of feature importance measures across the proposals and the literature methods considered. Results are reported as mean $\pm$ standard deviation over 50 runs. }\label{tab:importance}
\end{table}
Globally, Table~\ref{tab:importance} shows that the features with the highest proposed importance scores are $X_2$, $X_3$ and $X_1$, ranked in this order. 
As expected from the feature distribution, trimodal features ($X_2,X_3$) are the highest contributors in terms of importance scoring to $K=3$ partitions. Note that the proposed measure correctly penalizes the feature with highest intra-cluster variances ($X_3$) relative to the other ($X_2$), still assigning it a high score. Comparatively, the bimodal feature ($X_1$) is correctly assigned a lower score, being less informative than the previous two in discriminating between three groups. 
 All noisy features ($X_4,X_5,X_6$), instead, result in very low scores.
The results of the comparison with literature methods in terms of explainability show that
URF$_{\text{Gini}}$, URF$_{\text{perm}}$, CUBT, and LS obtain similar results, correctly identifying noisy variables but not being able to assign the correct importance ranking between $X_2$ and $X_3$. 
TWKM seems to have more difficulties in finding the right answers, even showing a great level of instability.

\subsubsection{Simulation study}
\label{sec:synthetic}

In this section, we design and simulate three datasets to provide a comprehensive evaluation of the proposed method, focusing on the impact of key feature characteristics such as cluster overlap, imbalanced cluster proportions, and feature correlation. In all cases, noisy features are added to further test robustness.\\
The data matrix $\mathbf{X}\in \mathbb{R}^{n\times (d_f+d_n)}$ is comprised by $d_f$ relevant features with different characteristics, and $d_n$ noisy features, with the following distributions: 
\begin{equation*}
    \begin{split}
        & X_{j}  \sim \left[ \sum_{k=1}^K p_k \ \mathcal{N}(\mu_k, \Sigma_k) \right]_j, \quad j=1,\dots,d_f ,\\
        & X_{j} \sim U([a_j,b_j]), \quad j = d_f+1, \dots, d_f+d_n,
    \end{split}
\end{equation*}
where, for the $k$th term in the  gaussian mixture: $p_k$ is the weight, $\mu_k$ is the $d_f$-dimensional mean, and $\Sigma_k$ is the $d_f \times d_f$-dimensional covariance matrix. 
The interval $[a_j, b_j]$ represents the uniform distribution's range.
Each component $k$ of the mixture represents a distinct cluster, and as a whole they represent the ground truth of $K$ clusters in terms of grouping of the dataset. Therefore, the modes of the gaussian components are chosen to be separated in all dimensions of the relevant features.  This controlled experimental setting allows a systematic assessment of the goal of our method, which is to provide an importance score relative to how informative each feature is \textit{in computing K-cluster partitions}. 
The parameters of the simulated datasets are summarized in Table~\ref{tab:parameters_synthetic}.
 The dimensions of variables are expressed as $d_{\cdot} = \sum_i (d_{\cdot})_i$, where, for relevant features, each $i$ corresponds to a block in the covariance matrix.
For noisy variables, each $i$ highlights the dimension of a group of features generated from the same marginal: a uniform with range sampled empirically from the  range of the blocks of features comprising the relevant ones: $[a_i,b_i] = [\min(X_{i_k}),\max(X_{i_k})]$, $i_k$ running over the possible indexes of block $i$.\\
More details about feature importance results are reported in Appendix \ref{appendix}.

\textbf{Overlap dataset.}
The results corresponding to the overlap setting in terms of clustering performances are summarized in Tables~\ref{tab:clustering_overlap_m_adjRand} and~\ref{tab:clustering_overlap_m_FM}. The cluster are well defined, leading to very high values of both the metrics for every clustering method. All the three settings of the proposed method improved the clusters definition with respect to the standard K-means.\\
In terms of feature importance, the results are summarized in Tables~\ref{tab:fi_overlap}, and~\ref{tab:fi_overlap_literature}. The proposed method clearly identifies the noisy variables for every setting and every choice of $m$, and the other variables are ranked accordingly to the ground truth of the data generation process: the first group $X_1,\dots,X_5$ have the maximum importance, the second one $X_6,\dots,X_{10}$ have lower values of importance due to the contribute of the higher variance, but they are still identified as very informative variables, while lower values of importance are assigned to the third group of informative variables $X_{11},\dots,,X_{15}$, generated with highest variance. \\
Comparing the results with the literature methods, URF$_{\text{Gini}}$, CUBT and LS result the most coherent with the data generation process, even if the differences between the scores are less extreme than using the proposed methods. It is not clear if TWKM is able to identify the noisy variables, since the differences in terms of importance scores are very small between the last block of informative variables and the noisy variables. All the other methods, except for URF$_{\text{perm}}$, correctly identify the noisy variables from the informative ones. \\

\textbf{Proportion dataset.}
The results corresponding to the imbalanced proportion scenario in terms of clustering performances are summarized in Tables~\ref{tab:clustering_proportion_m_adjRand} and~\ref{tab:clustering_proportion_m_FM}. The clustering definition is more difficult than the previous setting, leading to moderate values of both the metrics for the literature clustering methods. All three versions of the proposed method improved the clusters definition with respect to the literature methods, for every choice of $m$, although the Bayesian approach has a lower positive effect on results. \\
In terms of feature importance, the results are summarized in Tables~\ref{tab:fi_proportion} and~\ref{tab:fi_proportion_literature}. The proposed method clearly identifies the noisy variables for every setting and every choice of $m$, and the relevant variables are ranked accordingly to the ground truth of the data generation process: features in the first group $X_1,\dots, X_4$ have the highest importance, whereas features belonging to the second one $X_5,\dots,X_8$ have lower values of importance due to the higher variance in generation, albeit still identified as very informative. \\
Comparing the results with the literature methods, all of them are able to correctly rank the importance of the variables, with the exception of TWKM and URF$_{\text{perm}}$. However, some results are not completely coherent with the data generation process, e.g. CUBT overestimates the importance of noisy variables, showing instability.\\

\textbf{Correlation dataset.}
The results corresponding to the correlation setting in terms of clustering performances are summarized in Tables~\ref{tab:clustering_correlation_m_adjRand} and~\ref{tab:clustering_correlation_m_FM}. As in the Overlap setting, the clustering task is easy, leading to very high values of both the metrics for every clustering method. All the three versions of the proposed method improve the clusters definition with respect to the literature methods for every choice of $m$.\\
In terms of feature importance, the results are summarized in Tables~\ref{tab:fi_correlation} and~\ref{tab:fi_correlation_literature}. The proposed method clearly identifies the noisy variables for every setting and every choice of $m$, and the other variables are ranked accordingly to the ground truth of the data generation process, with some overlapping values: the first group $X_1,\dots,X_4$ have the maximum importance, the second one $X_5,\dots,X_8$ have lower values of importance due to the contribute of the higher variance, and lower values of importance are assigned to the third group of informative variables $X_{9},\dots,X_{12}$ that are generated with highest variance. The intra-cluster correlation does not affect the importance scoring. \\
Comparing the results with the literature methods, all the methods are able to correctly identify the noisy variables, but they are not always correct in the ranking assignments. For example TWKM assigns to $X_5$ a lower importance than $X_{12}$. Only CUBT correctly ranks the variables without overlapping cases.  \\

\begin{table}[h]
    \centering
    \renewcommand{\arraystretch}{1.8}
    \scalebox{0.85}{
    \begin{tabular}{lcccccc}
    \hline
         \textbf{Dataset} & $K$ & $N \cdot p$ & $d_f$ & $d_n$ & $\mu_k$ & $\Sigma_k$ \\
         \hline
         Overlap & 4 & (30,30,30,30) & 5+5+5 & 2+2+2 & $k\cdot  \mathbf{1}_{d_f}$ & 
         $\begin{pmatrix}
             (0.2)^2 \cdot \mathbf{1}_{5 \times 5} & \mathbf{0} & \mathbf{0} \\
             \mathbf{0} & (0.35)^2 \cdot \mathbf{1}_{5 \times 5} & \mathbf{0} \\
             \mathbf{0} & \mathbf{0} & (0.5)^2 \cdot \mathbf{1}_{5 \times 5} \\
         \end{pmatrix}$
         \\
         \hline
         Proportion & 5 & (15,20,40,30,15) & 4+4 & 4+4 & $k \cdot  \mathbf{1}_{d_f}$ & 
         $\begin{pmatrix}
             (0.25)^2 \cdot \mathbf{1}_{4 \times 4} & \mathbf{0} & \\
             \mathbf{0} & (0.5)^2 \cdot \mathbf{1}_{4 \times 4} \\
         \end{pmatrix}$
         \\
         \hline
Correlation 
& 4 
& (30,30,30,30) 
& 4+4+4 
& 2+2+2 
& $k \cdot \mathbf{1}_{d_f}$ 
& \makecell{$
\begin{pmatrix}
(0.25)^2 \mathbf{R} & \mathbf{0} & \mathbf{0} \\
\mathbf{0} & (0.35)^2 \mathbf{R} & \mathbf{0} \\
\mathbf{0} & \mathbf{0} & (0.45)^2 \mathbf{R}
\end{pmatrix}
$ \\ $\mathbf{R}_{i,j} = \lambda_i\lambda_j$, $\mathbf{R}_{i,i}=1$, $\lambda=(1,0.75,0.25,0.1)$}

\\
         \hline
          
    \end{tabular}
    }
    \caption{Dataset classes and specifications of the choice of parameters.}
\label{tab:parameters_synthetic}
\end{table}


\begin{table}[t]
\centering
\begin{subtable}
\linewidth
\begin{tabular}{ccccccc}
\hline
$m$ & \textbf{K-means} & \textbf{Hierarchical} & \textbf{Basic} & \textbf{Efron} & \textbf{BBC} \\
\hline
8  & $0.89 \pm 0.18$ & $1.00 \pm 0.00$ & $1.00 \pm 0.00$ & $1.00 \pm 0.00$ & $1.00 \pm 0.00$ \\
12 & $0.89 \pm 0.18$ & $1.00 \pm 0.00$ & $1.00 \pm 0.00$ & $1.00 \pm 0.00$ & $1.00 \pm 0.00$ \\
16 & $0.89 \pm 0.18$ & $1.00 \pm 0.00$ & $1.00 \pm 0.00$ & $1.00 \pm 0.00$ & $1.00 \pm 0.00$ \\
20 & $0.89 \pm 0.18$ & $1.00 \pm 0.00$ & $1.00 \pm 0.00$ & $1.00 \pm 0.00$ & $1.00 \pm 0.00$ \\
\hline
\end{tabular}
\caption{Adjusted Rand index.}
\label{tab:clustering_overlap_m_adjRand}
\end{subtable}
\begin{subtable}
\linewidth
\begin{tabular}{ccccccc}
\hline
$m$ & \textbf{K-means} & \textbf{Hierarchical} & \textbf{Basic} & \textbf{Efron} & \textbf{BBC} \\
\hline
8  & $0.92 \pm 0.12$ & $1.00 \pm 0.00$ & $1.00 \pm 0.00$ & $1.00 \pm 0.00$ & $1.00 \pm 0.00$ \\
12 & $0.92 \pm 0.12$ & $1.00 \pm 0.00$ & $1.00 \pm 0.00$ & $1.00 \pm 0.00$ & $1.00 \pm 0.00$ \\
16 & $0.92 \pm 0.12$ & $1.00 \pm 0.00$ & $1.00 \pm 0.00$ & $1.00 \pm 0.00$ & $1.00 \pm 0.00$ \\
20 & $0.92 \pm 0.12$ & $1.00 \pm 0.00$ & $1.00 \pm 0.00$ & $1.00 \pm 0.00$ & $1.00 \pm 0.00$ \\
\hline
\end{tabular}
\caption{Fowlkes-Mallows index.}
\label{tab:clustering_overlap_m_FM}
\end{subtable}
\caption{Clustering performance in terms of Adjusted Rand and Fowlkes-Mallows index of the overlap dataset. Results are shown as mean $\pm$ standard deviation over 20 replications for different values of the subspace size $m$. }
\label{tab:clustering_overlap_m}
\end{table}

\begin{table}[t]
\centering
\begin{subtable}{\textwidth}
\begin{tabular}{l l c c c c}
\hline
$m$ & \textbf{Method} & $X_1-X_5$ & $X_6-X_{10}$ & $X_{11}-X_{15}$ & $X_{16}-X_{21}$\\
  \hline
8 & Basic & $[0.99,\,1.00]$ & $[0.90,\,0.94]$ & $[0.73,\,0.83]$ & $[0.08,\,0.18]$ \\ 
  8 & Efron & $[0.99,\,1.00]$ & $[0.89,\,0.93]$ & $[0.72,\,0.82]$ & $[0.09,\,0.16]$ \\ 
  8 & BBC & $[0.99,\,1.00]$ & $[0.92,\,0.95]$ & $[0.74,\,0.85]$ & $[0.08,\,0.13]$ \\ 
  \hline
  12 & Basic & $[1.00,\,1.00]$ & $[0.95,\,0.97]$ & $[0.76,\,0.87]$ & $[0.01,\,0.05]$ \\ 
  12 & Efron & $[1.00,\,1.00]$ & $[0.92,\,0.95]$ & $[0.73,\,0.84]$ & $[0.03,\,0.05]$ \\ 
  12 & BBC & $[1.00,\,1.00]$ & $[0.95,\,0.97]$ & $[0.76,\,0.88]$ & $[0.01,\,0.05]$ \\ 
  \hline
  16 & Basic & $[1.00,\,1.00]$ & $[0.96,\,0.98]$ & $[0.76,\,0.88]$ & $[0.01,\,0.02]$ \\ 
  16 & Efron & $[1.00,\,1.00]$ & $[0.94,\,0.97]$ & $[0.74,\,0.86]$ & $[0.01,\,0.03]$ \\ 
  16 & BBC & $[1.00,\,1.00]$ & $[0.99,\,0.99]$ & $[0.79,\,0.92]$ & $[0.00,\,0.01]$ \\ 
  \hline
  20 & Basic & $[1.00,\,1.00]$ & $[0.96,\,0.98]$ & $[0.75,\,0.87]$ & $[0.00,\,0.02]$ \\ 
  20 & Efron & $[1.00,\,1.00]$ & $[0.95,\,0.98]$ & $[0.75,\,0.87]$ & $[0.01,\,0.02]$ \\ 
  20 & BBC & $[1.00,\,1.00]$ & $[1.00,\,1.00]$ & $[0.80,\,0.94]$ & $[0.00,\,0.00]$ \\  
   \hline
\end{tabular}%
\caption{Proposed feature importance scores calculated over 20 runs.  }
\label{tab:fi_overlap}
\end{subtable}

\begin{subtable}{\textwidth}
\begin{tabular}{l c c c c}
\hline
\textbf{Method} & $X_1-X_5$ & $X_6-X_{10}$ & $X_{11}-X_{15}$ & $X_{16}-X_{21}$\\
  \hline
  URF$_{\text{Gini}}$ & $[0.85,\,0.91]$ & $[0.71,\,0.77]$ & $[0.53,\,0.63]$ & $[0.01,\,0.04]$ \\ 
  URF$_{\text{perm}}$ & $[0.85,\,0.95]$ & $[0.85,\,0.85]$ & $[0.85,\,0.85]$ & $[0.85,\,0.85]$ \\ 
  CUBT & $[0.99,\,1.00]$ & $[0.91,\,0.94]$ & $[0.80,\,0.86]$ & $[0.12,\,0.17]$ \\ 
  TWKM & $[0.68,\,0.94]$ & $[0.10,\,0.25]$ & $[0.04,\,0.12]$ & $[0.01,\,0.07]$ \\ 
  LS & $[0.24,\,0.25]$ & $[0.29,\,0.30]$ & $[0.35,\,0.40]$ & $[0.96,\,1.00]$ \\  
   \hline
\end{tabular}%
\caption{Feature importance scores of literature methods calculated over 50 runs. }
\label{tab:fi_overlap_literature}
\end{subtable}
\caption{Feature importance scores for the overlap dataset. Results are expressed as $[\min,\max]$ of the average value per feature in each feature group. Features in the same group have same marginal distribution (see Table~\ref{tab:parameters_synthetic}).}\label{tab:fi_overlap}

\end{table}


\begin{table}[h]
\centering
\begin{subtable}
\linewidth
\begin{tabular}{ccccccc}
\hline
$m$ & \textbf{K-means} & \textbf{Hierarchical} & \textbf{Basic} & \textbf{Efron} & \textbf{BBC} \\
\hline
9  & $0.52 \pm 0.11$ & $0.61 \pm 0.00$ & $0.72 \pm 0.09$ & $0.73 \pm 0.11$ & $0.63 \pm 0.08$ \\
11 & $0.52 \pm 0.11$ & $0.61 \pm 0.00$ & $0.71 \pm 0.10$ & $0.69 \pm 0.07$ & $0.65 \pm 0.11$ \\
13 & $0.52 \pm 0.11$ & $0.61 \pm 0.00$ & $0.73 \pm 0.07$ & $0.72 \pm 0.05$ & $0.66 \pm 0.07$ \\
15 & $0.52 \pm 0.11$ & $0.61 \pm 0.00$ & $0.75 \pm 0.11$ & $0.72 \pm 0.04$ & $0.64 \pm 0.05$ \\
\hline
\end{tabular}
\caption{Adjusted Rand index.}
\label{tab:clustering_proportion_m_adjRand}
\end{subtable}
\begin{subtable}
\linewidth
\begin{tabular}{ccccccc}
\hline
$m$ & \textbf{K-means} & \textbf{Hierarchical} & \textbf{Basic} & \textbf{Efron} & \textbf{BBC} \\
\hline
9  & $0.62 \pm 0.09$ & $0.69 \pm 0.00$ & $0.79 \pm 0.07$ & $0.80 \pm 0.08$ & $0.72 \pm 0.06$ \\
11 & $0.62 \pm 0.09$ & $0.69 \pm 0.00$ & $0.78 \pm 0.08$ & $0.77 \pm 0.05$ & $0.74 \pm 0.08$ \\
13 & $0.62 \pm 0.09$ & $0.69 \pm 0.00$ & $0.80 \pm 0.06$ & $0.79 \pm 0.04$ & $0.75 \pm 0.06$ \\
15 & $0.62 \pm 0.09$ & $0.69 \pm 0.00$ & $0.81 \pm 0.08$ & $0.79 \pm 0.03$ & $0.74 \pm 0.03$ \\
\hline
\end{tabular}
\caption{Fowlkes-Mallows index.}
\label{tab:clustering_proportion_m_FM}
\end{subtable}
\caption{Clustering performance in terms of Adjusted Rand and Fowlkes-Mallows index of the proportion dataset. Results are shown as mean $\pm$ standard deviation over 20 replications for different values of the subspace size $m$.}
\label{tab:clustering_proportion_m}
\end{table}

\begin{table}[t]
\centering
\begin{subtable}{\linewidth}
\begin{tabular}{l l c c c c}
\hline
$m$ & \textbf{Method} & $X_1-X_4$ & $X_5-X_{8}$ & $X_{9}-X_{16}$ \\
 \hline
9 & Basic & $[0.95,\,0.98]$ & $[0.74,\,0.88]$ & $[0.11,\,0.22]$ \\ 
  9 & Efron & $[0.96,\,0.98]$ & $[0.75,\,0.87]$ & $[0.12,\,0.22]$ \\ 
  9 & BBC & $[0.96,\,0.98]$ & $[0.75,\,0.87]$ & $[0.13,\,0.23]$ \\ 
  \hline
  11 & Basic & $[0.95,\,0.99]$ & $[0.70,\,0.87]$ & $[0.06,\,0.15]$ \\ 
  11 & Efron & $[0.96,\,0.98]$ & $[0.72,\,0.86]$ & $[0.07,\,0.15]$ \\ 
  11 & BBC & $[0.96,\,0.98]$ & $[0.72,\,0.85]$ & $[0.07,\,0.16]$ \\ 
  \hline
  13 & Basic & $[0.95,\,0.99]$ & $[0.69,\,0.87]$ & $[0.03,\,0.12]$ \\ 
  13 & Efron & $[0.95,\,0.98]$ & $[0.69,\,0.85]$ & $[0.04,\,0.12]$ \\ 
  13 & BBC & $[0.95,\,0.98]$ & $[0.69,\,0.84]$ & $[0.04,\,0.12]$ \\
  \hline
  15 & Basic & $[0.96,\,0.99]$ & $[0.69,\,0.88]$ & $[0.02,\,0.09]$ \\ 
  15 & Efron & $[0.96,\,0.98]$ & $[0.68,\,0.85]$ & $[0.03,\,0.09]$ \\ 
  15 & BBC & $[0.95,\,0.98]$ & $[0.68,\,0.83]$ & $[0.03,\,0.09]$ \\ 
   \hline
\end{tabular}%
\caption{Proposed feature importance scores calculated over 20 runs. }
\label{tab:fi_proportion}
\end{subtable}
\begin{subtable}{\linewidth}
\begin{tabular}{l l c c c c}
\hline
\textbf{Method} & $X_1-X_4$ & $X_5-X_{8}$ & $X_{9}-X_{16}$ \\
 \hline
URF$_{\text{Gini}}$ & $[0.85,\,0.91]$ & $[0.59,\,0.76]$ & $[0.03,\,0.07]$ \\ 
URF$_{\text{perm}}$  & $[0.50,\,0.72]$ & $[0.05,\,0.30]$ & $[0.00,\,0.00]$ \\ 
CUBT & $[0.98,\,1.00]$ & $[0.84,\,0.88]$ & $[0.15,\,0.31]$ \\ 
TWKM & $[0.68,\,0.82]$ & $[0.15,\,0.35]$ & $[0.04,\,0.14]$ \\ 
LS & $[0.50,\,0.51]$ & $[0.55,\,0.61]$ & $[0.98,\,1.00]$ \\ 
   \hline
\end{tabular}%
\caption{Feature importance scores of literature methods calculated over 50 runs. }
\label{tab:fi_proportion_literature}
\end{subtable}
\caption{Feature importance scores for the proportion dataset. Results are expressed as $[\min,\max]$ of the average value per feature in each feature group. Features in the same group have same marginal distribution (see Table~\ref{tab:parameters_synthetic}).}\label{tab:fi_proportion}
\end{table}

\begin{table}[t]
\centering
\begin{subtable}{\linewidth}
\begin{tabular}{ccccccc}
\hline
$m$ & \textbf{K-means} & \textbf{Hierarchical} & \textbf{Basic} & \textbf{Efron} & \textbf{BBC} \\
\hline
9  & $0.89 \pm 0.20$ & $0.91 \pm 0.00$ & $1.00 \pm 0.00$ & $1.00 \pm 0.01$ & $1.00 \pm 0.01$ \\
11 & $0.89 \pm 0.20$ & $0.91 \pm 0.00$ & $1.00 \pm 0.00$ & $1.00 \pm 0.01$ & $1.00 \pm 0.00$ \\
13 & $0.89 \pm 0.20$ & $0.91 \pm 0.00$ & $1.00 \pm 0.01$ & $1.00 \pm 0.00$ & $1.00 \pm 0.00$ \\
15 & $0.89 \pm 0.20$ & $0.91 \pm 0.00$ & $1.00 \pm 0.00$ & $1.00 \pm 0.00$ & $1.00 \pm 0.01$ \\
\hline
\end{tabular}
\caption{Adjusted Rand index of the correlation dataset.}
\label{tab:clustering_correlation_m_adjRand}
\end{subtable}

\begin{subtable}{\linewidth}
\centering
\begin{tabular}{ccccccc}
\hline
 $m$ & \textbf{K-means} & \textbf{Hierarchical} & \textbf{Basic} & \textbf{Efron} & \textbf{BBC} \\
\hline
9  & $0.92 \pm 0.15$ & $0.94 \pm 0.00$ & $1.00 \pm 0.00$ & $1.00 \pm 0.00$ & $1.00 \pm 0.00$ \\
11 & $0.92 \pm 0.15$ & $0.94 \pm 0.00$ & $1.00 \pm 0.00$ & $1.00 \pm 0.00$ & $1.00 \pm 0.00$ \\
13 & $0.92 \pm 0.15$ & $0.94 \pm 0.00$ & $1.00 \pm 0.01$ & $1.00 \pm 0.00$ & $1.00 \pm 0.00$ \\
15 & $0.92 \pm 0.15$ & $0.94 \pm 0.00$ & $1.00 \pm 0.00$ & $1.00 \pm 0.00$ & $1.00 \pm 0.00$ \\
\hline
\end{tabular}
\caption{Fowlkes-Mallows index of the correlation dataset.}
\label{tab:clustering_correlation_m_FM}
\end{subtable}
\caption{Clustering performance in terms of Adjusted Rand and Fowlkes-Mallows index of the correlation dataset. Results are shown as mean $\pm$ standard deviation over 20 replications for different values of the subspace size $m$.}
\label{tab:clustering_correlation_m}
\end{table}

\begin{table}[t]
\centering
\begin{subtable}{\linewidth}
\begin{tabular}{l l c c c c}
\hline
$m$ & \textbf{Method} & $X_1-X_4$ & $X_5-X_{8}$ & $X_{9}-X_{12}$ & $X_{13}-X_{18}$\\
\hline
9 & Basic & $[0.97,\,1.00]$ & $[0.93,\,0.96]$ & $[0.86,\,0.90]$ & $[0.06,\,0.12]$ \\ 
  9 & Efron & $[0.96,\,0.99]$ & $[0.92,\,0.94]$ & $[0.84,\,0.88]$ & $[0.08,\,0.11]$ \\ 
  9 & BBC & $[0.97,\,1.00]$ & $[0.93,\,0.96]$ & $[0.86,\,0.90]$ & $[0.07,\,0.10]$ \\ 
  \hline
  11 & Basic & $[0.98,\,1.00]$ & $[0.95,\,0.97]$ & $[0.87,\,0.92]$ & $[0.02,\,0.06]$ \\ 
  11 & Efron & $[0.97,\,1.00]$ & $[0.92,\,0.95]$ & $[0.83,\,0.88]$ & $[0.04,\,0.07]$ \\ 
  11 & BBC & $[0.97,\,1.00]$ & $[0.93,\,0.96]$ & $[0.85,\,0.90]$ & $[0.04,\,0.06]$ \\ 
  \hline
  13 & Basic & $[0.98,\,1.00]$ & $[0.96,\,0.98]$ & $[0.89,\,0.94]$ & $[0.01,\,0.04]$ \\ 
  13 & Efron & $[0.97,\,1.00]$ & $[0.93,\,0.96]$ & $[0.84,\,0.90]$ & $[0.02,\,0.05]$ \\ 
  13 & BBC & $[0.98,\,1.00]$ & $[0.95,\,0.97]$ & $[0.86,\,0.92]$ & $[0.02,\,0.04]$ \\ 
  \hline
  15 & Basic & $[0.99,\,1.00]$ & $[0.97,\,0.98]$ & $[0.91,\,0.96]$ & $[0.01,\,0.04]$ \\ 
  15 & Efron & $[0.98,\,1.00]$ & $[0.94,\,0.96]$ & $[0.85,\,0.91]$ & $[0.02,\,0.04]$ \\ 
  15 & BBC & $[0.99,\,1.00]$ & $[0.96,\,0.98]$ & $[0.88,\,0.94]$ & $[0.01,\,0.04]$ \\ 
   \hline

\end{tabular}%
\caption{Proposed feature importance scores calculated on 20 runs. Results are reported for different subspace size $m$.}
\label{tab:fi_correlation}
\end{subtable}
\begin{subtable}{\linewidth}
    \begin{tabular}{l c c c c}
\hline
\textbf{Method} & $X_1-X_4$ & $X_5-X_{8}$ & $X_{9}-X_{12}$ & $X_{13}-X_{18}$\\
  \hline
  URF$_{\text{Gini}}$ & $[0.82,\,0.90]$ & $[0.78,\,0.82]$ & $[0.63,\,0.81]$ & $[0.01,\,0.03]$ \\ 
  URF$_{\text{perm}}$ & $[0.13,\,0.32]$ & $[0.16,\,0.28]$ & $[0.06,\,0.18]$ & $[0.05,\,0.05]$ \\ 
  CUBT & $[0.97,\,1.00]$ & $[0.93,\,0.95]$ & $[0.87,\,0.91]$ & $[0.13,\,0.22]$ \\ 
  TWKM & $[0.57,\,0.86]$ & $[0.19,\,0.40]$ & $[0.09,\,0.23]$ & $[0.01,\,0.02]$ \\ 
  LS & $[0.28,\,0.31]$ & $[0.30,\,0.32]$ & $[0.33,\,0.37]$ & $[0.95,\,1.00]$ \\  
   \hline
\end{tabular}%
\caption{Feature importance scores of literature methods calculated on 50 runs. }
\label{tab:fi_correlation_literature}
\end{subtable}
\caption{Feature importance scores for the correlation dataset. Results are expressed as $[\min,\max]$ of the average value of each feature in each feature group. Features in the same group have same marginal distribution (see Table~\ref{tab:parameters_synthetic}).}\label{tab_fi_correlation}
\end{table}

\clearpage

\subsection{Real dataset}

\subsubsection{Wine}
\begin{figure}[ht!]
    \centering
    \begin{subfigure}[t]{0.48\textwidth}
        \centering
        \includegraphics[width=\linewidth]{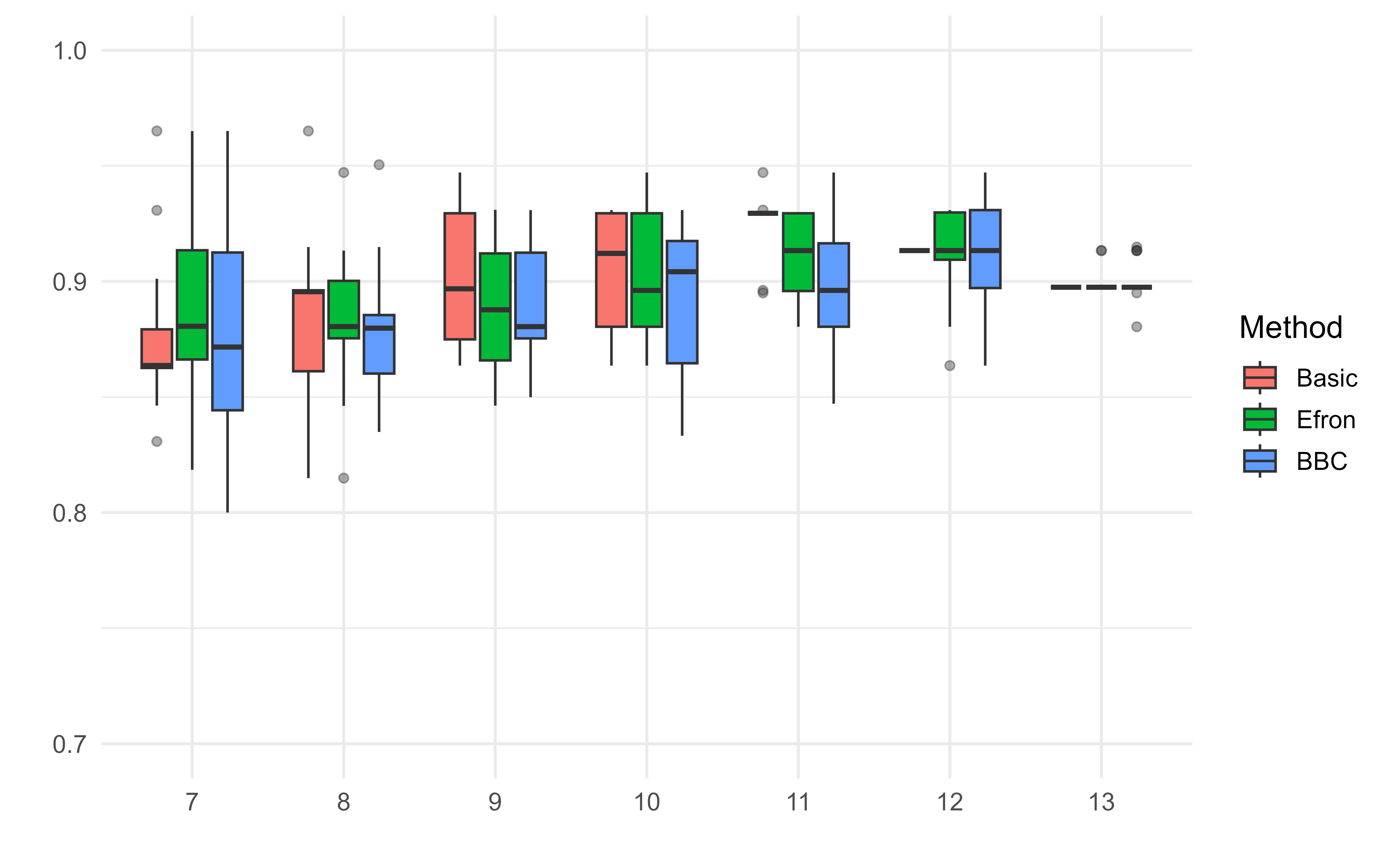}
        \caption{Adjusted Rand index}
        \label{fig:adjRand_wine_per_m}
    \end{subfigure}\hfill
    \begin{subfigure}[t]{0.48\textwidth}
        \centering
        \includegraphics[width=\linewidth]{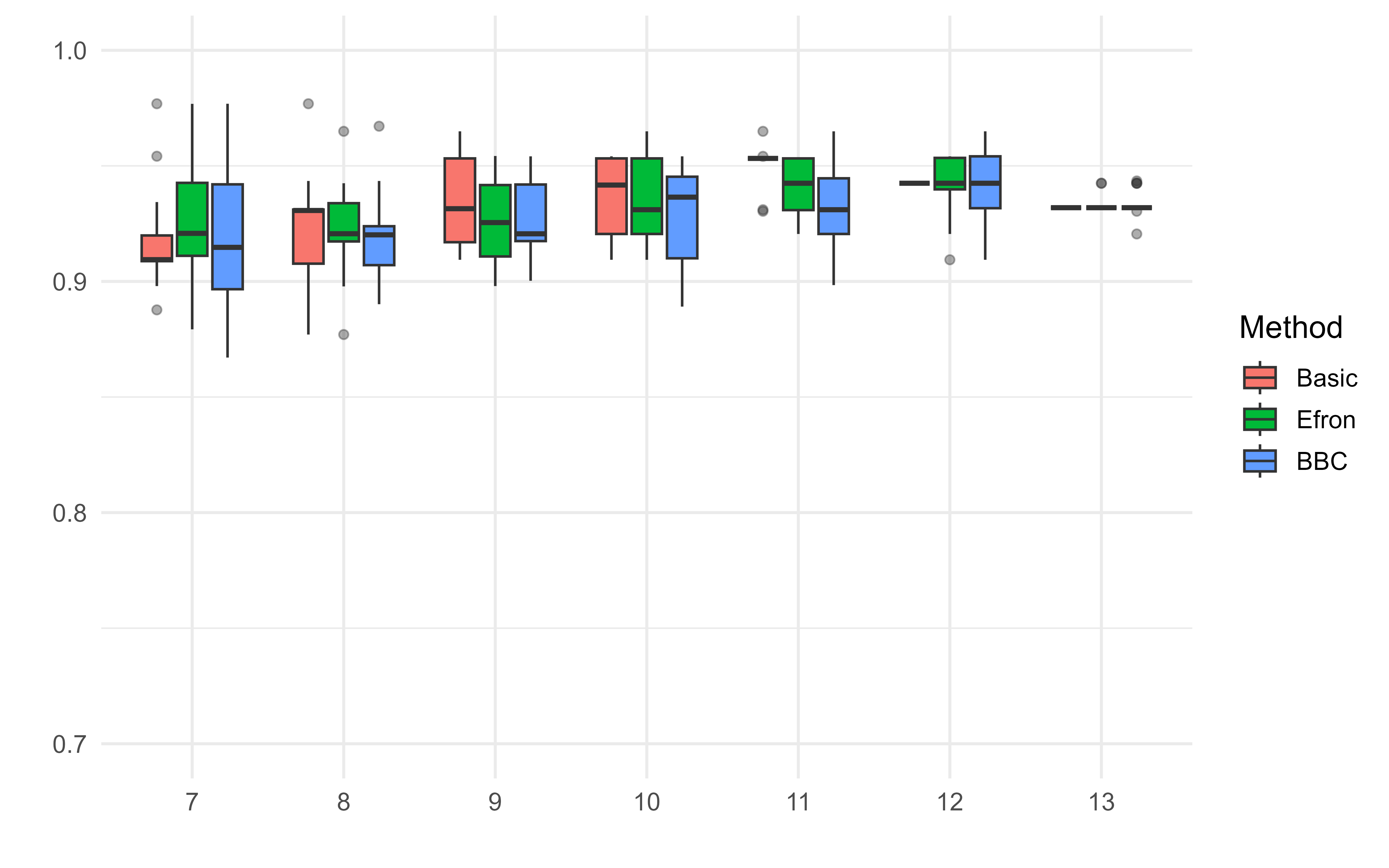}
        \caption{Fowlkes-Mallows index}
        \label{fig:FM_wine_per_m}
    \end{subfigure}
\caption{Clustering performance of the proposed method across the three resampling schemes on the Wine dataset. Results are computed in terms of Adjusted Rand and Fowlkes-Mallows index, and averaged over 20 runs on the dataset, for different subspace size choices. The horizontal axis represents the subspace size chosen ($m$).}
    \label{fig:clustering_wine_per_m}
\end{figure}

\begin{figure}[ht!]
    \centering    
    \includegraphics[width=\linewidth]{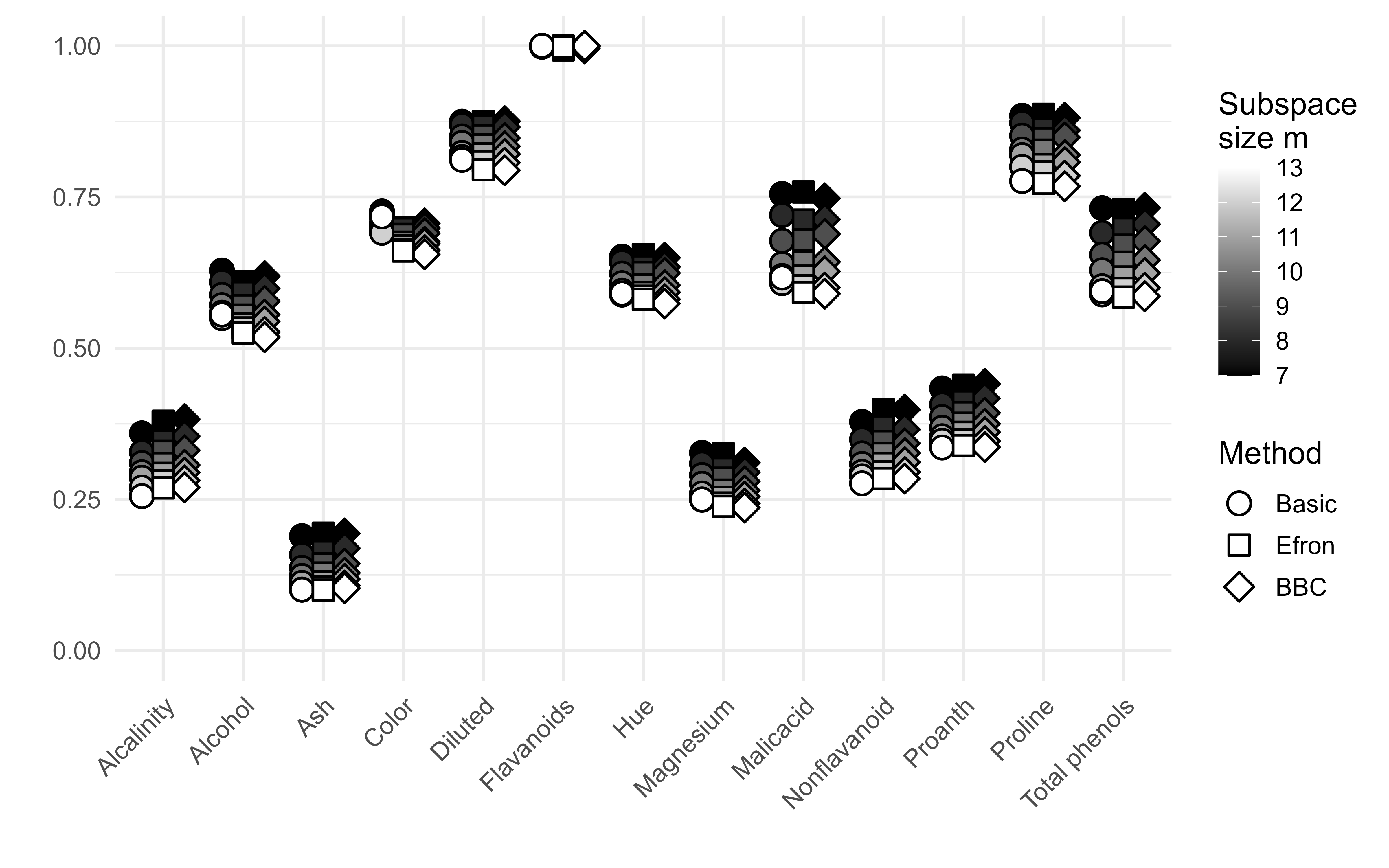}
    \caption{Comparison of the mean importance of each feature evaluated by our three proposals for different choices of subspace size, over 20 runs on the Wine dataset.}
    \label{fig:MeanVariableImportance_wine_per_m}
\end{figure}

As a first real example, we consider the \textit{Wine dataset} from the UCI Machine Learning Repository \cite{wine_109}. It is a widely used benchmark dataset for classification and pattern recognition tasks, containing 178 samples of wines derived from three different cultivars. Each sample is characterized by 13 features representing continuous physicochemical attributes: alcalinity of ash (Alcalinity), alcohol content (Alcohol), ash, color intensity (Color), flavonoids, hue,  magnesium, malic acid (Malicacid), non-flavanoid phenols (Nonflavonoid), proanthocyanins (Proanth), proline concentration (Proline), total phenols, and 0D280 0D315 of diluted wines (Diluted).\\
We consider as ground truth cluster label the categorical class identifying the wine cultivar ($K=3$). The dataset is notable for its relatively low noise, and separated classes. Additionally, the moderate correlations among its features make the dataset a good test case for feature selection. 
In terms of clustering performances, see Table~\ref{tab:clustering_wine} and Figure~\ref{fig:clustering_wine_per_m}, the structure of the data does not adapt to the hierarchical method, that achieves moderate values in both the evaluation metrics. The proposed methods slightly improve the results of K-means for the dropout parameter choice $m=p-1$. \\
Varying $m$ the results remain stable, confirming the robustness of the method in a real setting, especially for moderate to low levels of dropout of features ($m=9,\dots,12$).
The stability of the proposed feature importance score with respect to $m$ is investigated in Figure~\ref{fig:MeanVariableImportance_wine_per_m}. Although numerical differences are observed, they have a negligible impact on the ranking. The results are extremely consistent across the three approaches proposed.\\
The feature importance scores obtained are compared to the literature methods in Tables~\ref{tab:wine_fi_comparison} and~\ref{tab:wine_feature_scores}.
Proposed scores, at fixed $m=12$, are consistent across the three resampling variants, which lead to similar values and the same ranking of variable importance. All of the methods, with the exception of TWKM, agree on assigning the highest importance to the Flavonoids variable.
All methods broadly agree on the least important features, while the importance of all other features show disagreements between methods, e.g. Diluted and Proline are ranked as 2nd and 3rd by the proposed methods, while the literature methods put them between the first and the fifth positions. 

\begin{table}[ht]
\centering

\begin{tabular}{lccccc}
\toprule
Metric & K-means & Hierarchical & Basic & Efron & BBC \\
\midrule
Adjusted Rand
& $0.90 \pm 0$ & $0.58 \pm 0$ & $0.91 \pm 0$ & $0.91 \pm 0.02$ & $0.91 \pm 0.02$ \\

Fowlkes--Mallows
& $0.93 \pm 0$ & $0.72 \pm 0$ & $0.94 \pm 0$ & $0.94 \pm 0.01$ & $0.94 \pm 0.01$ \\
\bottomrule
\end{tabular}
\caption{Clustering performances are reported as mean $\pm$ standard deviation over 50 runs of each method on the Wine dataset. The subspace size is set as $m=12$.}
\label{tab:clustering_wine}
\end{table}


\begin{table}[ht]
\centering
\begin{subtable}{\linewidth}
\centering
\resizebox{\textwidth}{!}{
\begin{tabular}
{lccccccccccccc}
\toprule
\textbf{Method} &
Alcohol &
Malicacid &
Ash &
Alcalinity &
Magnesium &
Total phenols &
Flavanoids &
Nonflavanoid &
Proanth &
Color&
Hue &
Diluted &
Proline \\
\midrule
Basic
& $0.55 \pm 0.01$ & $0.61 \pm 0.01$ & $0.10 \pm 0$ & $0.27 \pm 0$
& $0.25 \pm 0$ & $0.59 \pm 0$ & $1.00 \pm 0$ & $0.29 \pm 0$
& $0.35 \pm 0$ & $0.69 \pm 0.01$ & $0.59 \pm 0.01$
& $0.82 \pm 0.01$ & $0.80 \pm 0.01$ \\

Efron
& $0.53 \pm 0.01$ & $0.60 \pm 0.01$ & $0.11 \pm 0$ & $0.28 \pm 0$
& $0.25 \pm 0.01$ & $0.60 \pm 0.01$ & $1.00 \pm 0$ & $0.30 \pm 0.01$
& $0.35 \pm 0$ & $0.66 \pm 0.01$ & $0.59 \pm 0.01$
& $0.81 \pm 0.01$ & $0.79 \pm 0.01$ \\

BBC
& $0.53 \pm 0.01$ & $0.60 \pm 0.01$ & $0.11 \pm 0$ & $0.28 \pm 0$
& $0.24 \pm 0.01$ & $0.60 \pm 0.01$ & $1.00 \pm 0$ & $0.30 \pm 0$
& $0.35 \pm 0$ & $0.66 \pm 0.01$ & $0.58 \pm 0.01$
& $0.80 \pm 0.01$ & $0.78 \pm 0.01$ \\
\hline
ranking &8& 5/6& 13 & 11 &12 & 5/6&1 &10 & 9& 4& 7& 2 & 3 \\
\bottomrule
\end{tabular}
}
\caption{Proposed feature importance scores for the Wine dataset. The subspace size is set as $m=12$.}
\label{tab:wine_fi_comparison}
\end{subtable}
\begin{subtable}{\linewidth}
\centering
\resizebox{\textwidth}{!}{
\begin{tabular}{lccccccccccccc}
\toprule
\textbf{Method} &
Alcohol &
Malicacid &
Ash &
Alcalinity &
Magnesium &
Total phenols &
Flavanoids &
Nonflavanoid &
Proanth &
Color &
Hue &
Diluted &
Proline \\
\midrule
URF$_{\text{Gini}}$ &
$0.33\!\pm\!0.07$ & $0.16\!\pm\!0.05$ & $0.03\!\pm\!0.03$ & $0.08\!\pm\!0.06$ &
$0.02\!\pm\!0.03$ & $0.65\!\pm\!0.13$ & $1.00\!\pm\!0.00$ & $0.09\!\pm\!0.05$ &
$0.34\!\pm\!0.10$ & $0.42\!\pm\!0.06$ & $0.33\!\pm\!0.07$ & $0.63\!\pm\!0.08$ &
$0.48\!\pm\!0.11$ \\
 &7/8& 9& 12 & 11 &13 & 2&1 &10 & 6& 5& 7/8& 3& 4 \\

URF$_{\text{perm}}$ &
$0.05\!\pm\!0.05$ & $0.00\!\pm\!0.01$ & $0.00\!\pm\!0.00$ & $0.01\!\pm\!0.01$ &
$0.00\!\pm\!0.00$ & $0.47\!\pm\!0.23$ & $0.99\!\pm\!0.02$ & $0.00\!\pm\!0.01$ &
$0.04\!\pm\!0.04$ & $0.15\!\pm\!0.07$ & $0.06\!\pm\!0.05$ & $0.48\!\pm\!0.16$ &
$0.13\!\pm\!0.08$ \\
 &7& 10/13& 10/13 & 9 &10/13 & 3&1 &10/13 & 8& 4& 6& 2& 5 \\
CUBT &
$0.88\!\pm\!0.00$ & $0.39\!\pm\!0.00$ & $0.48\!\pm\!0.00$ & $0.29\!\pm\!0.00$ &
$0.61\!\pm\!0.00$ & $0.81\!\pm\!0.00$ & $1.00\!\pm\!0.00$ & $0.21\!\pm\!0.00$ &
$0.61\!\pm\!0.00$ & $0.63\!\pm\!0.00$ & $0.68\!\pm\!0.00$ & $0.80\!\pm\!0.00$ &
$0.96\!\pm\!0.00$ \\
 &3& 11& 10 & 12 &8/9 & 4&1 &13 & 8/9& 7& 6& 5& 2 \\
TWKM &
$0.10\!\pm\!0.29$ & $0.40\!\pm\!0.49$ & $0.10\!\pm\!0.30$ & $0.00\!\pm\!0.02$ &
$0.00\!\pm\!0.00$ & $0.00\!\pm\!0.01$ & $0.34\!\pm\!0.45$ & $0.11\!\pm\!0.30$ &
$0.05\!\pm\!0.22$ & $0.26\!\pm\!0.43$ & $0.01\!\pm\!0.02$ & $0.31\!\pm\!0.45$ &
$1.00\!\pm\!0.00$ \\
&7/8& 2& 7/8 & 11/13&11/13 & 11/13&3 &6 & 9& 5& 10& 4& 1 \\
LS &
$0.78\!\pm\!0.00$ & $0.85\!\pm\!0.00$ & $1.00\!\pm\!0.00$ & $0.86\!\pm\!0.00$ &
$0.94\!\pm\!0.00$ & $0.63\!\pm\!0.00$ & $0.50\!\pm\!0.00$ & $0.81\!\pm\!0.00$ &
$0.82\!\pm\!0.00$ & $0.77\!\pm\!0.00$ & $0.75\!\pm\!0.00$ & $0.62\!\pm\!0.00$ &
$0.68\!\pm\!0.00$ \\
&7& 10& 13 & 11 &12 & 3&1 &8 & 9& 6& 5& 2& 4 \\
\bottomrule
\end{tabular}
}
\caption{Feature importance scores of literature methods for the Wine dataset. }
\label{tab:wine_feature_scores}
\end{subtable}\caption{Feature importance scores for the Wine dataset. Results are reported as mean $\pm$ standard deviation over 50 runs. Rankings of features for each method are reported.}
\label{tab:wine_fi}
\end{table}

\subsubsection{Breast Cancer Original Dataset}

\begin{figure}[ht!]
    \centering
    \begin{subfigure}[t]{0.48\textwidth}
        \centering
        \includegraphics[width=\linewidth]{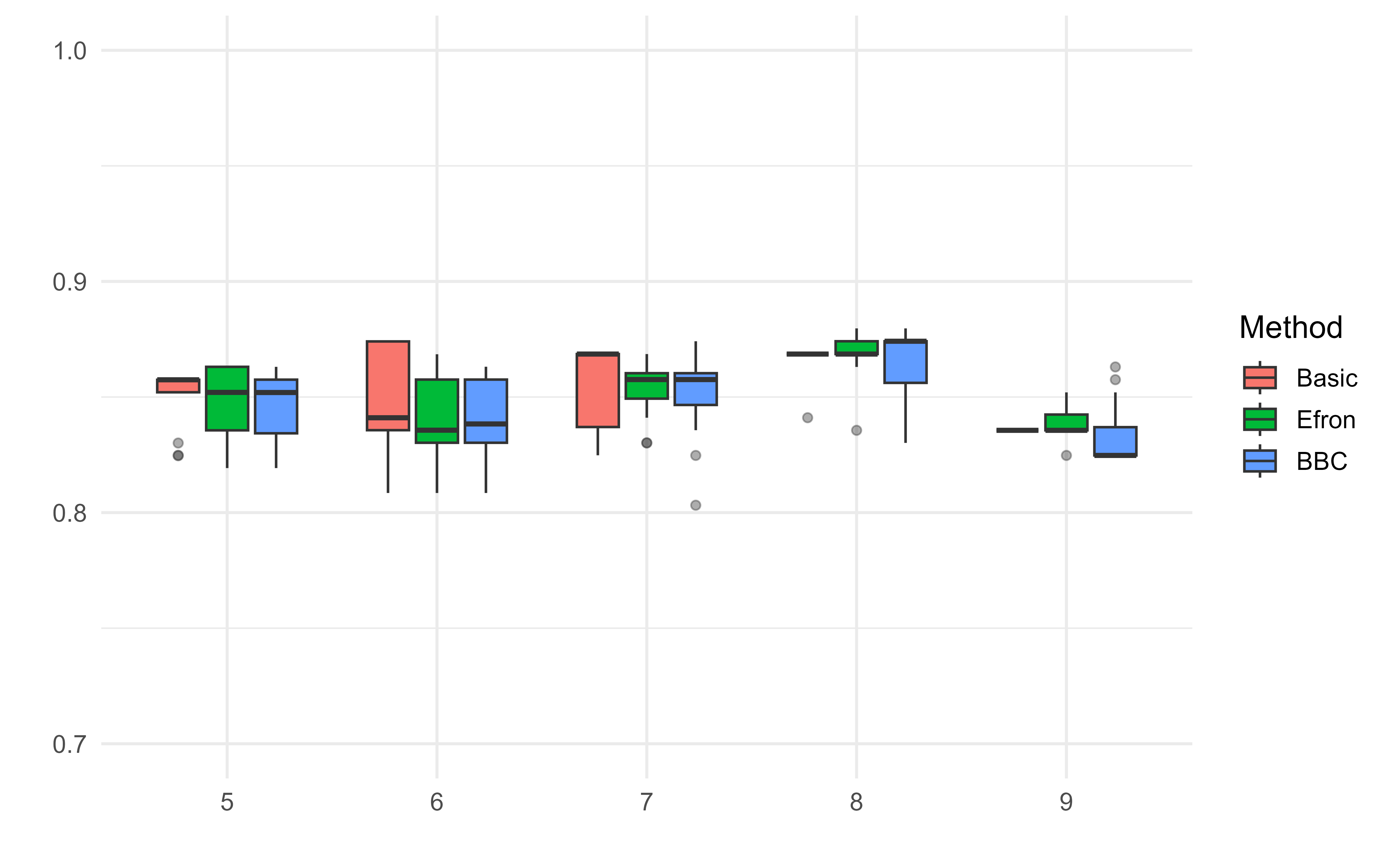}
        \caption{Adjusted Rand index}
        \label{fig:adjRand_breast_per_m}
    \end{subfigure}\hfill
    \begin{subfigure}[t]{0.48\textwidth}
        \centering
        \includegraphics[width=\linewidth]{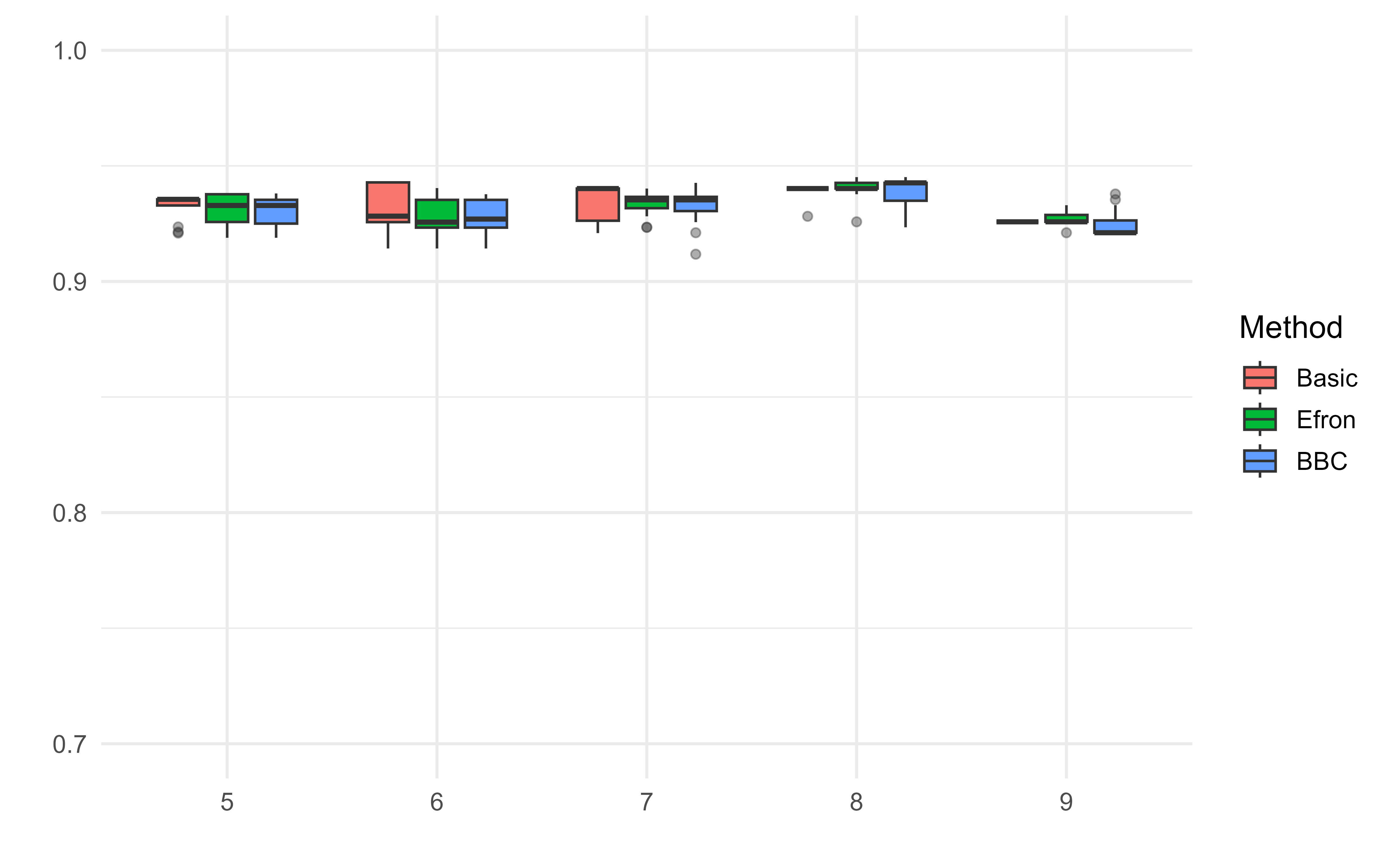}
        \caption{Fowlkes-Mallows index}
        \label{fig:FM_breast_per_m}
    \end{subfigure}
\caption{Clustering performance of the proposed method across the three resampling schemes on the Breast dataset. Results are computed in terms of Adjusted Rand and Fowlkes-Mallows index, and averaged over 50 runs on the dataset, for different subspace size choices. The horizontal axis represents the subspace size chosen ($m$).}\label{fig:clustering_breast_per_m}
\end{figure}

\begin{figure}[ht!]
    \centering    
    \includegraphics[width=\linewidth]{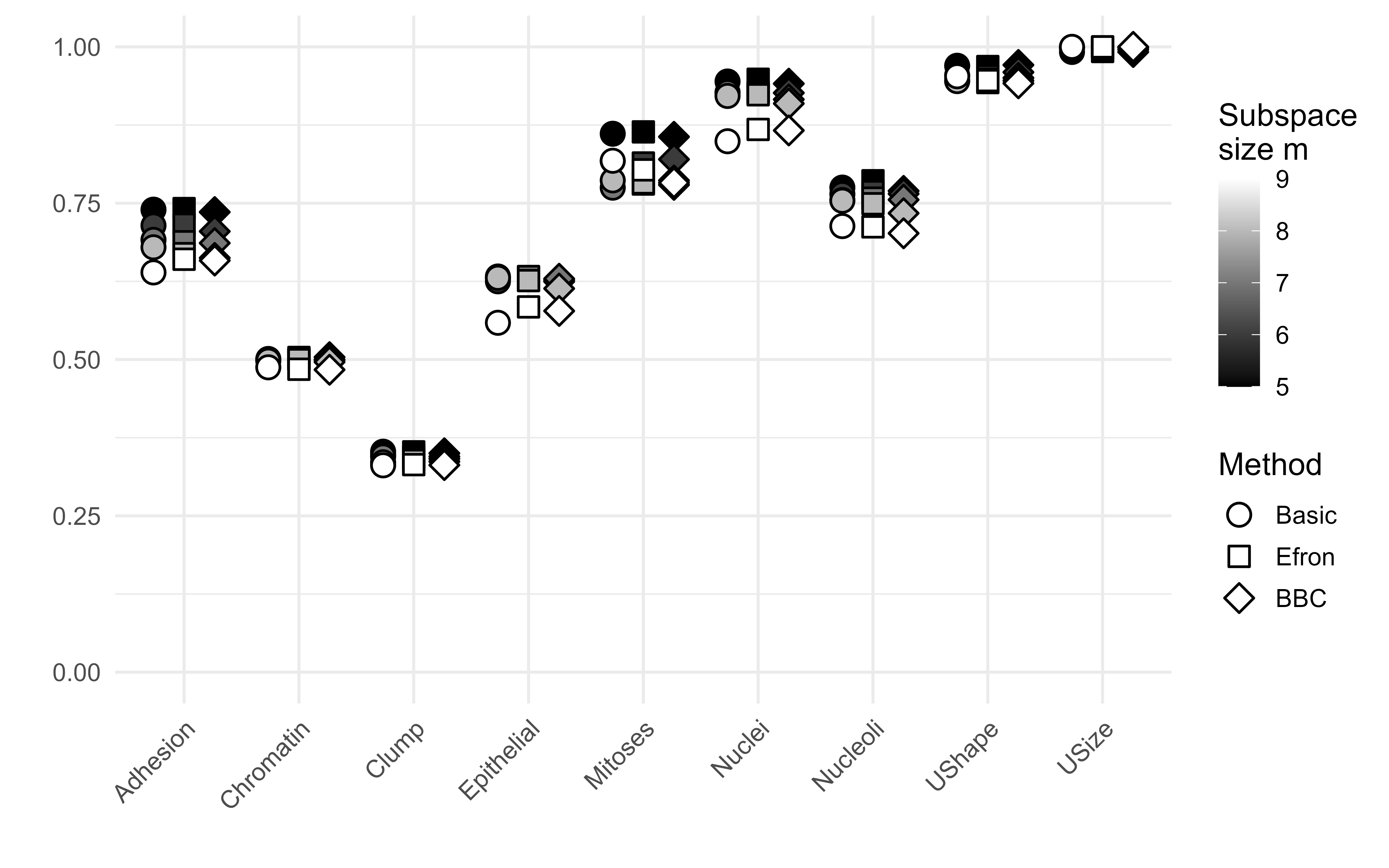}
    \caption{Comparison of the mean importance of each feature evaluated by the three versions of the proposal for different choices of subspace size $m$, over 20 runs on the Breast dataset.}
    \label{fig:MeanVariableImportance_breast_per_m}
\end{figure}

As a second real example, we consider the \textit{Breast Cancer Wisconsin (Original) dataset} from the UCI Machine Learning Repository \cite{Wolberg1990Breast}. It is another benchmark dataset for classification and clustering tasks \cite{wolberg1990multisurface}, containing $699$ samples described by $9$ integer-valued cytological features. 
The features capture morphological characteristics of the cell nuclei, including clump thickness (Clump), uniformity of cell size (USize), uniformity of cell shape (UShape), marginal adhesion (Adhesion), single epithelial cell size (Epithelial), bare nuclei (Nuclei), bland chromatin (Chromatin), normal nucleoli (Nucleoli), and mitoses (Mitoses).
We consider as ground truth cluster label the binary class identifying the diagnosis, distinguishing malignant cases (34$\%$ prevalence) from benign ones. 
\\
As for the previous real example, for varying $m$ the results remain stable, confirming the robustness of the method in a real setting, especially for moderate to low levels of dropout ($m=7,8$), see Figure~\ref{fig:clustering_breast_per_m}. Comparing the results for $m=8$ to hierarchical clustering and K-means, see Table \ref{tab:clustering_breast}, a similar behaviour of the previous real example is observed. Hierarchical clustering achieve moderate values of performance metrics, and the three variants of the proposed metrics slightly improve the classical K-means approach.
The stability of the proposed feature importance score with respect to $m$ is investigated in Figure~\ref{fig:MeanVariableImportance_breast_per_m}. Although numerical differences are observed, they have a negligible impact on the ranking. The results are extremely consistent across the three approaches proposed.\\ The feature importance scores obtained are compared to the literature methods in Tables~\ref{tab:fi_comparison_breast} and~\ref{tab:feature_comparison_breast}. The proposed approach are consistent, leading to extremely similar scores and the same ranking of variable importances. All methods agree on assigning the highest importance to the USize variable and the second rank is assigned to UShape by all the methods except TWKM, that places the variable as the least important. The importance of the other variables show fundamental disagreements between methods. URF$_{\text{Gini}}$ , URF$_{\text{perm}}$  and CUBT seem consistent in the ranking, showing some differences with respect to the proposed methods, e.g they assign to Mitosis very low scores, while the proposed approaches put the variable as the 4th most important one, and an opposite behavior is showed for the Chromatin variable, in agreement with TWKM and LS.   

\begin{table}[ht]
\centering
\begin{tabular}{lccccc}
\toprule
Metric & K-means & Hierarchical & Basic & Efron & BBC \\
\midrule
Adjusted Rand
& $0.84 \pm 0$ & $0.47 \pm 0$ & $0.87 \pm 0.01$ & $0.87 \pm 0.01$ & $0.86 \pm 0.02$ \\

Fowlkes--Mallows
& $0.93 \pm 0$ & $0.79 \pm 0$ & $0.94 \pm 0$ & $0.94 \pm 0$ & $0.94 \pm 0.01$ \\
\bottomrule
\end{tabular}
\caption{Clustering performances are reported as mean $\pm$ standard deviation over 50 runs of each method on the Breast dataset. The subspace size of the proposed method is set as $m=8$.}
\label{tab:clustering_breast}
\end{table}

\begin{table}[ht]
\centering
\begin{subtable}{\linewidth}
\centering
\resizebox{\textwidth}{!}{
\begin{tabular}{lccccccccc}
\toprule
\textbf{Method} &
Clump & USize & UShape & Adhesion & Epithelial & Nuclei & Chromatin & Nucleoli & Mitoses \\
\midrule
FI Basic
& $0.34 \pm 0$ & $1.00 \pm 0$ & $0.95 \pm 0$ & $0.68 \pm 0$
& $0.63 \pm 0.01$ & $0.92 \pm 0.01$ & $0.50 \pm 0$
& $0.76 \pm 0.01$ & $0.79 \pm 0.01$ \\

FI Efron
& $0.34 \pm 0$ & $1.00 \pm 0$ & $0.94 \pm 0$ & $0.67 \pm 0$
& $0.63 \pm 0.01$ & $0.93 \pm 0.01$ & $0.50 \pm 0$
& $0.75 \pm 0.01$ & $0.78 \pm 0.01$ \\

FI BBC
& $0.34 \pm 0$ & $1.00 \pm 0$ & $0.95 \pm 0$ & $0.66 \pm 0$
& $0.61 \pm 0.01$ & $0.91 \pm 0.01$ & $0.50 \pm 0$
& $0.74 \pm 0.01$ & $0.78 \pm 0.01$ \\
\hline
Ranking
& 9 & 1 & 2 & 6 & 7 & 3 & 8 & 5 & 4 \\
\bottomrule
\end{tabular}
}
\caption{Proposed feature importance scores for the Breast dataset. The subspace size is chosen as $m=8$.}
\label{tab:fi_comparison_breast}
\end{subtable}
\begin{subtable}{\linewidth}
\centering
\resizebox{\textwidth}{!}{
\begin{tabular}{lccccccccc}
\toprule
\textbf{Method} &
Clump & USize & UShape & Adhesion & Epithelial & Nuclei & Chromatin & Nucleoli & Mitoses \\
\midrule
URF$_{\text{Gini}}$ &
$0.48\!\pm\!0.03$ & $0.99\!\pm\!0.02$ & $0.92\!\pm\!0.08$ & $0.41\!\pm\!0.05$ &
$0.57\!\pm\!0.08$ & $0.58\!\pm\!0.07$ & $0.60\!\pm\!0.05$ & $0.43\!\pm\!0.06$ &
$0.00\!\pm\!0.00$ \\
& 6 & 1 & 2 & 8 & 5 & 4 & 3 & 7 & 9 \\

URF$_{\text{perm}}$ &
$0.37\!\pm\!0.04$ & $0.99\!\pm\!0.02$ & $0.91\!\pm\!0.06$ & $0.33\!\pm\!0.06$ &
$0.52\!\pm\!0.10$ & $0.48\!\pm\!0.05$ & $0.55\!\pm\!0.06$ & $0.36\!\pm\!0.07$ &
$0.00\!\pm\!0.00$ \\
& 6 & 1 & 2 & 8 & 4 & 5 & 3 & 7 & 9 \\

CUBT &
$0.68\!\pm\!0.00$ & $1.00\!\pm\!0.00$ & $0.97\!\pm\!0.00$ & $0.83\!\pm\!0.00$ &
$0.92\!\pm\!0.00$ & $0.83\!\pm\!0.00$ & $0.92\!\pm\!0.00$ & $0.83\!\pm\!0.00$ &
$0.64\!\pm\!0.00$ \\
& 8 & 1 & 2 & 4/5/6 & 3 & 4/5/6 & 3 & 4/5/6 & 9 \\

TWKM &
$0.20\!\pm\!0.40$ & $0.45\!\pm\!0.50$ & $0.00\!\pm\!0.00$ & $0.15\!\pm\!0.36$ &
$0.05\!\pm\!0.22$ & $0.10\!\pm\!0.30$ & $0.10\!\pm\!0.30$ & $0.15\!\pm\!0.36$ &
$0.35\!\pm\!0.48$ \\
& 3 & 1 & 9 & 4/5 & 8 & 6/7 & 6/7 & 4/5 & 2 \\

LS &
$1.00\!\pm\!0.00$ & $0.65\!\pm\!0.00$ & $0.70\!\pm\!0.00$ & $0.81\!\pm\!0.00$ &
$0.84\!\pm\!0.00$ & $0.78\!\pm\!0.00$ & $0.85\!\pm\!0.00$ & $0.78\!\pm\!0.00$ &
$0.99\!\pm\!0.00$ \\
& 9 & 1 & 2 & 6 & 7 & 4/5 & 8 & 4/5 & 3 \\
\bottomrule
\end{tabular}
}
\caption{Feature importance scores of literature methods for the Breast dataset.}
\label{tab:feature_comparison_breast}
\end{subtable} \caption{Feature importance scores for the Breast dataset. Results are reported as mean $\pm $ standard deviation over 50 runs. Rankings of features for each method are reported. }\label{tab:fi_breast}
\end{table}

\section{Conclusions and future research} \label{sec:conclusions}

This work proposes a methodology able to assess explainability in cluster analysis, extending previous literature by combining a rich set of theoretical and applied techniques. Bagged clustering, internal validation and information theory are employed for consensus clustering and accurate estimation of feature importance. \\
Empirical results on simulated datasets show that the proposed method not only recovers cluster structures accurately, leveraging bagging to enhance existing techniques, but also produces feature importance metrics that align coherently with the true data-generating process.
Results on benchmark real datasets confirm the applicability of the methodology in more complex scenarios.\\
Overall, the empirical results demonstrate that the proposed model improves current approaches by providing stable cluster assignments along with a coherent feature importance score. In the comparison with approaches in the literature, emerges the importance of methodologies capable of jointly ensuring a high-quality cluster definition and consistent variable importance measure. Within such a framework, both the cluster construction process as well as the contribution of each variable can be interpreted in a unified perspective. \\
The complexity in time of the proposal is based on that detailed in \cite{quetti2025bayesian} for the BBC procedure, with the addition of the estimation of entropies. Therefore, the complexity from being linear becomes quadratic in the number of data points due to the KDE step. \\ 
Future perspectives point to different directions. In the proposal, estimation of the number of clusters $K$ can be implemented through BBC method \cite{quetti2025bayesian}, other optimal $K$ selection approaches (e.g. silhouette \cite{silhouette}), or cluster algorithms that do not need a predefined $K$  (e.g. DBSCAN \cite{dbscan} or other density based approaches). Other consensus clustering approaches can be considered \cite{ghosh2011cluster}, depending on the specific context of application. Finally, our explainability framework could be combined with measures of accuracy and robustness for unsupervised learning, based on the mutual information, in line with the safe machine learning framework proposed in \cite{auricchio2025rank, babaei2025}.
We remark that all steps of the proposed procedure can be reproduced using the R functions and data available in a GitHub repository.

\section*{Acknowledgements}
F.M.Q. acknowledges RES C.E.O. Federico Bonelli for the award of Ph.D. scholarship.

\bibliography{main_bibliography}

@inproceedings{Lundberg2017,
  author    = {Lundberg, S. M. and Lee, S.},
  title     = {A unified approach to interpreting model predictions},
  booktitle = {Adv. Neural Inf. Process. Syst.},
  year      = {2017},
  volume    = {30},
  pages     = {4768--4777}
}

@article{babaei2025,
  author  = {Babaei, G. and Giudici, P. and Raffinetti, E.},
  title   = {A rank graduation box for safe artificial intelligence},
  journal = {Expert Syst. Appl.},
  year    = {2025}
}

@article{breiman1996bagging,
  title     = {Bagging predictors},
  author    = {Breiman, Leo},
  journal   = {Machine Learning},
  volume    = {24},
  number    = {2},
  pages     = {123--140},
  year      = {1996},
  publisher = {Springer}
}

@article{gokcay2002information,
  title={Information theoretic clustering},
  author={Gokcay, Erhan and Principe, Jose C.},
  journal={IEEE transactions on pattern analysis and machine intelligence},
  volume={24},
  number={2},
  pages={158--171},
  year={2002},
  publisher={IEEE}
}

@article{solorio2020review,
  title={A review of unsupervised feature selection methods},
  author={Solorio-Fern{\'a}ndez, Sa{\'u}l and Carrasco-Ochoa, J Ariel and Mart{\'\i}nez-Trinidad, Jos{\'e} Fco},
  journal={Artificial Intelligence Review},
  volume={53},
  number={2},
  pages={907--948},
  year={2020},
  publisher={Springer}
}

@inproceedings{faivishevsky2012unsupervised,
  title={Unsupervised feature selection based on non-parametric mutual information},
  author={Faivishevsky, Lev and Goldberger, Jacob},
  booktitle={2012 IEEE International Workshop on Machine Learning for Signal Processing},
  pages={1--6},
  year={2012},
  organization={IEEE}
}

@article{franccois2007resampling,
  title={Resampling methods for parameter-free and robust feature selection with mutual information},
  author={Fran{\c{c}}ois, Damien and Rossi, Fabrice and Wertz, Vincent and Verleysen, Michel},
  journal={Neurocomputing},
  volume={70},
  number={7-9},
  pages={1276--1288},
  year={2007},
  publisher={Elsevier}
}

@article{torkkola2003feature,
  title={Feature extraction by non-parametric mutual information maximization},
  author={Torkkola, Kari},
  journal={Journal of machine learning research},
  volume={3},
  number={Mar},
  pages={1415--1438},
  year={2003}
}

@article{kozachenko1987sample,
  title={Sample estimate of the entropy of a random vector},
  author={Kozachenko, Leonenko},
  journal={Probl. Pered. Inform.},
  volume={23},
  pages={9},
  year={1987}
}

@incollection{davis2011remarks,
  title={Remarks on some nonparametric estimates of a density function},
  author={Davis, Richard A and Lii, Keh-Shin and Politis, Dimitris N},
  booktitle={Selected Works of Murray Rosenblatt},
  pages={95--100},
  year={2011},
  publisher={Springer},
  address={New York, NY}
}

@article{parzen1962estimation,
  title={On estimation of a probability density function and mode},
  author={Parzen, Emanuel},
  journal={The annals of mathematical statistics},
  volume={33},
  number={3},
  pages={1065--1076},
  year={1962},
  publisher={JSTOR}
}

@article{dy2004feature,
  title={Feature selection for unsupervised learning},
  author={Dy, Jennifer G and Brodley, Carla E},
  journal={Journal of machine learning research},
  volume={5},
  number={Aug},
  pages={845--889},
  year={2004}
}

@article{liu2009feature,
  title={Feature selection with dynamic mutual information},
  author={Liu, Huawen and Sun, Jigui and Liu, Lei and Zhang, Huijie},
  journal={Pattern Recognition},
  volume={42},
  number={7},
  pages={1330--1339},
  year={2009},
  publisher={Elsevier}
}

@article{vergara2014review,
  title={A review of feature selection methods based on mutual information},
  author={Vergara, Jorge R and Est{\'e}vez, Pablo A},
  journal={Neural computing and applications},
  volume={24},
  pages={175--186},
  year={2014},
  publisher={Springer}
}

@book{cover1999elements,
  title={Elements of information theory},
  author={Cover, Thomas M},
  year={1999},
  publisher={John Wiley \& Sons},
    address   = {Hoboken, NJ}
}

@article{shannon1948mathematical,
  title={A mathematical theory of communication},
  author={Shannon, Claude E},
  journal={The Bell system technical journal},
  volume={27},
  number={3},
  pages={379--423},
  year={1948},
  publisher={Nokia Bell Labs}
}

@article{chen2011tw,
  title={TW-k-means: Automated two-level variable weighting clustering algorithm for multiview data},
  author={Chen, Xiaojun and Xu, Xiaofei and Huang, Joshua Zhexue and Ye, Yunming},
  journal={IEEE Transactions on Knowledge and Data Engineering},
  volume={25},
  number={4},
  pages={932--944},
  year={2011},
  publisher={IEEE}
}

@article{importanceCUBT,
  title={Assessing variable importance in clustering: a new method based on unsupervised binary decision trees},
  author={Badih, G. and Pierre, M. and Laurent, B.},
  journal={Computational Statistics},
  volume={34},
  pages={301--321},
  year={2019},
  doi={10.1007/s00180-018-0857-0}
}

@article{breiman2001random,
  title={Random forests},
  author={Breiman, Leo},
  journal={Machine learning},
  volume={45},
  number={1},
  pages={5--32},
  year={2001},
  publisher={Springer}
}

@article{belkin2001laplacian,
  title={Laplacian eigenmaps and spectral techniques for embedding and clustering},
  author={Belkin, Mikhail and Niyogi, Partha},
  journal={Advances in neural information processing systems},
  volume={14},
  year={2001}
}

@inproceedings{jmac1967,
  author		= "James MacQueen",
  title			= "Some methods for classification and analysis of multivariate observations.",
  booktitle		= "Proceedings of the fifth Berkeley symposium on mathematical statistics and probability",
  volume="1",
  number="14",
  pages="281-297",
  year			= "1967"
}

@article{dbscan,
  author		= "Ester, M. and Kriegel, H. P. and Sander, J. and Xu, X.",
  title			= "A density-based algorithm for discovering clusters in large spatial databases with noise.",
  journal		= "kdd",
  volume		= "96",
  number      =  "34",
  pages			= "226-231",
  year			= "1996",
  doi                = ""
}

@article{hierarchical,
  author		= "Murtagh, F. and Contreras, P.",
  title			= "Algorithms for hierarchical clustering: an overview.",
  journal		= "Wiley Interdisciplinary Reviews: Data Mining and Knowledge Discovery",
  volume		= "2",
  number      =  "1",
  pages			= "86-97",
  year			= "2012",
  doi                = ""
}

@book{Elements2009,
  title     = "The elements of statistical learning: data mining, inference, and prediction",
  author    = "Hastie, T. and Tibshirani, R. and Friedman, J. H.",
  year      = "2009",
  volume="2",
  pages="1-758",
  publisher = "Springer",
  address   = "New York"
}

@article{jain1999data, 
title="Data clustering: a review",
  author="Jain, Anil K and Murty, M Narasimha and Flynn, Patrick J",
  journal="ACM computing surveys (CSUR)",
  volume="31",
  number="3",
  pages="264--323",
  year="1999",
  publisher="Acm New York, NY, USA"
}

@article{efron1992bootstrap,
author = {B. Efron},
title = {{Bootstrap Methods: Another Look at the Jackknife}},
volume = {7},
journal = {The Annals of Statistics},
number = {1},
publisher = {Institute of Mathematical Statistics},
pages = {1 -- 26},
year = {1979},
doi = {10.1214/aos/1176344552},
}

@article{rubin1981bayesian,
  title="The bayesian bootstrap",
  author="Rubin, Donald B",
  journal="The annals of statistics",
  pages="130--134",
  year="1981",
  publisher="JSTOR"
}

@article{Lo1987,
  author		= "Lo, A. Y.",
  title			= "A large sample study of the Bayesian bootstrap.",
  journal		= "The Annals of Statistics",
  volume		= "15",
  number      =  "1",
  pages			= "360-375",
  year			= "1987",
  doi                = "10.1214/aos/1176350271"
}

@article{galvani2021, 
title="A Bayesian nonparametric learning approach to ensemble models using the proper Bayesian bootstrap",
  author="Galvani, Marta and Bardelli, Chiara and Figini, Silvia and Muliere, Pietro",
  journal="Algorithms",
  volume="14",
  number="1",
  pages="11",
  year="2021",
  publisher="MDPI"
}

@article{muliere1996bayesian,
  title="Bayesian nonparametric predictive inference and bootstrap techniques",
  author="Muliere, Pietro and Secchi, Piercesare",
  journal="Annals of the institute of statistical mathematics",
  volume="48",
  pages="663--673",
  year="1996",
  publisher="Springer"
}

@article{dudoit2003bagging,
  title="Bagging to improve the accuracy of a clustering procedure",
  author="Dudoit, Sandrine and Fridlyand, Jane",
  journal="Bioinformatics",
  volume="19",
  number="9",
  pages="1090--1099",
  year="2003",
  publisher="Oxford University Press"
}

@techreport{Leisch,
 title = "Bagged clustering",
author = "Friedrich Leisch",
year = "1999",
doi = "10.57938/9b129f95-b53b-44ce-a129-5b7a1168d832",
series = "Working Papers SFB. Adaptive Information Systems and Modelling in Economics and Management Science",
number = "51",
publisher = "SFB Adaptive Information Systems and Modelling in Economics and Management Science, WU Vienna University of Economics and Business",
edition = "August 1999",
type = "WorkingPaper",
institution = "SFB Adaptive Information Systems and Modelling in Economics and Management Science, WU Vienna University of Economics and Business",
}

@article{interpretClust,
      title="Interpretable Clustering: A Survey", 
      author="Lianyu Hu and Mudi Jiang and Junjie Dong and Xinying Liu and Zengyou He",
      year="2024",
      eprint="2409.00743",
      archivePrefix="arXiv",
      primaryClass="cs.LG",
      url="https://arxiv.org/abs/2409.00743", 
}

@article{validation,
title = {From A-to-Z review of clustering validation indices},
journal = {Neurocomputing},
volume = {601},
pages = {128198},
year = {2024},
doi = {https://doi.org/10.1016/j.neucom.2024.128198},
author = {Bryar A. Hassan and Noor Bahjat Tayfor and Alla A. Hassan and Aram M. Ahmed and Tarik A. Rashid and Naz N. Abdalla}
}

@article{CUBT,
 title="Interpretable clustering
using unsupervised binary trees", 
      author="R. Fraiman and B. Ghattas and M. Svarc",
     journal = "Advances in Data Analysis and Classification",
volume = "7",
pages = "125-145",
year = "2013",
doi = "https://doi.org/10.1007/s11634-013-0129-3",
}

@article{ruleClust,
 title="On clustering and interpreting with rules by means of mathematical optimization", 
      author="E. Carrizosa and K. Kurishchenko and A. Mar´ın and D. Romero Morales",
     journal = "Computers \& Operations Research",
volume = "154",
pages = "106180",
year = "2023",
doi = "https://doi.org/10.1016/j.cor.2023.106180",
}

@article{quetti2025bayesian,
  title={A Bayesian approach to ensemble clustering},
  author={Quetti, Federico Maria and Figini, Silvia and Ballante, Elena},
  journal={Statistics},
  pages={1--23},
  year={2025},
  publisher={Taylor \& Francis}
}

@article{jing2007entropy,
  title={An entropy weighting k-means algorithm for subspace clustering of high-dimensional sparse data},
  author={Jing, Liping and Ng, Michael K and Huang, Joshua Zhexue},
  journal={IEEE Transactions on knowledge and data engineering},
  volume={19},
  number={8},
  pages={1026--1041},
  year={2007},
  publisher={IEEE}
}

@article{URF,
author = {Mantero, Alejandro and Ishwaran, Hemant},
title = {Unsupervised random forests},
journal = {Statistical Analysis and Data Mining: The ASA Data Science Journal},
volume = {14},
number = {2},
pages = {144-167},
doi = {https://doi.org/10.1002/sam.11498},
year = {2021}
}

@article{auricchio2025rank,
  title={On Rank Graduation Metrics for High Dimensional Ordinal Data},
  author={Auricchio, Gennaro and Bernardelli, Adelaide Emma and Giudici, Paolo and Toscani, Giuseppe},
  journal={arXiv preprint arXiv:2511.23100},
  year={2025}
}

@article{wade2018bayesian,
  title={Bayesian cluster analysis: Point estimation and credible balls (with discussion)},
  author={Wade, Sara and Ghahramani, Zoubin},
  year={2018}
}

@article{jain2015asymptotic,
  title={Asymptotic Behavior of Mean Partitions in Consensus Clustering},
  author={Jain, Brijnesh},
  journal={arXiv preprint arXiv:1512.06061},
  year={2015}
}

@inproceedings{meilua2003comparing,
  title={Comparing clusterings by the variation of information},
  author={Meil{\u{a}}, Marina},
  booktitle={Learning Theory and Kernel Machines: 16th Annual Conference on Learning Theory and 7th Kernel Workshop, COLT/Kernel 2003, Washington, DC, USA, August 24-27, 2003. Proceedings},
  pages={173--187},
  year={2003},
  organization={Springer}
}

@article{monti2003consensus,
  title={Consensus clustering: a resampling-based method for class discovery and visualization of gene expression microarray data},
  author={Monti, Stefano and Tamayo, Pablo and Mesirov, Jill and Golub, Todd},
  journal={Machine learning},
  volume={52},
  number={1},
  pages={91--118},
  year={2003},
  publisher={Springer}
}

@article{hubert1985comparing,
  title={Comparing partitions},
  author={Hubert, Lawrence and Arabie, Phipps},
  journal={Journal of Classification},
  volume={2},
  number={1},
  pages={193--218},
  year={1985}
}

@misc{wine_109,
  author       = {Aeberhard, Stefan and Forina, M.},
  title        = {{Wine}},
  year         = {1992},
  howpublished = {UCI Machine Learning Repository},
  doi         = {https://doi.org/10.24432/C5PC7J}
}

@article{ghosh2011cluster,
  title={Cluster ensembles},
  author={Ghosh, Joydeep and Acharya, Ayan},
  journal={Wiley interdisciplinary reviews: Data mining and knowledge discovery},
  volume={1},
  number={4},
  pages={305--315},
  year={2011},
  publisher={Wiley Online Library},
}

@inproceedings{nguyen2007consensus,
  title={Consensus clusterings},
  author={Nguyen, Nam and Caruana, Rich},
  booktitle={Seventh IEEE international conference on data mining (ICDM 2007)},
  pages={607--612},
  year={2007},
  organization={IEEE}
}

@inproceedings{dhillon2004kernel,
  title={Kernel k-means: spectral clustering and normalized cuts},
  author={Dhillon, Inderjit S and Guan, Yuqiang and Kulis, Brian},
  booktitle={Proceedings of the tenth ACM SIGKDD international conference on Knowledge discovery and data mining},
  pages={551--556},
  year={2004}
}

@article{lloyd1982least,
  title={Least squares quantization in PCM},
  author={Lloyd, Stuart},
  journal={IEEE transactions on information theory},
  volume={28},
  number={2},
  pages={129--137},
  year={1982},
  publisher={IEEE}
}

@article{ng2001spectral,
  title={On spectral clustering: Analysis and an algorithm},
  author={Ng, Andrew and Jordan, Michael and Weiss, Yair},
  journal={Advances in neural information processing systems},
  volume={14},
  year={2001}
}

@book{mclachlan2000finite,
  title={Finite mixture models},
  author={McLachlan, Geoffrey J and Peel, David},
  year={2000},
  publisher={John Wiley \& Sons},
    address   = {Hoboken, NJ}
}

@article{nino2021feature,
  title={Feature weighting methods: A review},
  author={Ni{\~n}o-Adan, Iratxe and Manjarres, Diana and Landa-Torres, Itziar and Portillo, Eva},
  journal={Expert Systems with Applications},
  volume={184},
  pages={115424},
  year={2021},
  publisher={Elsevier}
}

@article{oskouei2025feature,
  title={Feature-weighted fuzzy clustering methods: An experimental review},
  author={Oskouei, Amin Golzari and Samadi, Negin and Khezri, Shirin and Moghaddam, Arezou Najafi and Babaei, Hamidreza and Hamini, Kiavash and Nojavan, Saghar Fath and Bouyer, Asgarali and Arasteh, Bahman},
  journal={Neurocomputing},
  volume={619},
  pages={129176},
  year={2025},
  publisher={Elsevier}
}

@article{rand1971objective,
  title={Objective criteria for the evaluation of clustering methods},
  author={Rand, William M},
  journal={Journal of the American Statistical association},
  volume={66},
  number={336},
  pages={846--850},
  year={1971},
  publisher={Taylor \& Francis}
}

@misc{Wolberg1990Breast,
  author       = {Wolberg, WIlliam},
  title        = {{Breast Cancer Wisconsin (Original)}},
  year         = {1990},
  howpublished = {UCI Machine Learning Repository},
  doi         = {https://doi.org/10.24432/C5HP4Z}
}

@article{wolberg1990multisurface,
  title={Multisurface method of pattern separation for medical diagnosis applied to breast cytology.},
  author={Wolberg, William H and Mangasarian, Olvi L},
  journal={Proceedings of the national academy of sciences},
  volume={87},
  number={23},
  pages={9193--9196},
  year={1990}
}

@article{ferguson1973bayesian,
  title="A Bayesian analysis of some nonparametric problems",
  author="Ferguson, Thomas S",
  journal="The annals of statistics",
  pages="209--230",
  year="1973",
  publisher="JSTOR"
}

@article{fowlkes1983method,
  title={A method for comparing two hierarchical clusterings},
  author={Fowlkes, Edward B and Mallows, Colin L},
  journal={Journal of the American statistical association},
  volume={78},
  number={383},
  pages={553--569},
  year={1983},
  publisher={Taylor \& Francis}
}

@article{silhouette,
  author		= "Rousseeuw, Peter J. ",
  title			= "Silhouettes: a graphical aid to the interpretation and validation of cluster analysis.",
  journal		= "Journal of computational and applied mathematics",
  volume		= "20",
  pages			= "53-65",
  year			= "1987",
  doi                = ""
}

\clearpage

\appendix
\section{Appendix}
\label{appendix}

We report the full tables of results for the synthetic datasets in Section~\ref{sec:synthetic}.

\begin{sidewaystable}[t]
\centering

\begin{subtable}{\textwidth}
\centering

\resizebox{\textwidth}{!}{%
\begin{tabular}{cl|ccccc|ccccc|ccccc|cccccc}
\hline
$m$ & \textbf{Method} &
$X_1$ & $X_2$ & $X_3$ & $X_4$ & $X_5$ & $X_6$ & $X_7$ & $X_8$ & $X_9$ & $X_{10}$ & $X_{11}$ & $X_{12}$ & $X_{13}$ & $X_{14}$ & $X_{15}$ & $X_{16}$ & $X_{17}$ & $X_{18}$ & $X_{19}$ & $X_{20}$ & $X_{21}$ \\ 
  \hline
8 & Basic & $1.00 \pm 0.00$ & $0.99 \pm 0.00$ & $0.99 \pm 0.00$ & $1.00 \pm 0.00$ & $1.00 \pm 0.00$ & $0.94 \pm 0.01$ & $0.93 \pm 0.01$ & $0.90 \pm 0.01$ & $0.93 \pm 0.01$ & $0.91 \pm 0.02$ & $0.83 \pm 0.02$ & $0.81 \pm 0.02$ & $0.76 \pm 0.01$ & $0.73 \pm 0.02$ & $0.78 \pm 0.01$ & $0.18 \pm 0.02$ & $0.11 \pm 0.02$ & $0.08 \pm 0.01$ & $0.12 \pm 0.02$ & $0.12 \pm 0.02$ & $0.12 \pm 0.03$ \\ 
  8 & Efron & $1.00 \pm 0.00$ & $0.99 \pm 0.00$ & $0.99 \pm 0.00$ & $1.00 \pm 0.00$ & $0.99 \pm 0.00$ & $0.93 \pm 0.01$ & $0.92 \pm 0.01$ & $0.89 \pm 0.02$ & $0.91 \pm 0.01$ & $0.90 \pm 0.01$ & $0.82 \pm 0.03$ & $0.80 \pm 0.02$ & $0.76 \pm 0.01$ & $0.72 \pm 0.02$ & $0.78 \pm 0.01$ & $0.16 \pm 0.03$ & $0.11 \pm 0.02$ & $0.09 \pm 0.02$ & $0.12 \pm 0.02$ & $0.13 \pm 0.02$ & $0.14 \pm 0.03$ \\ 
  8 & BBC & $1.00 \pm 0.00$ & $0.99 \pm 0.00$ & $0.99 \pm 0.00$ & $1.00 \pm 0.00$ & $1.00 \pm 0.00$ & $0.95 \pm 0.01$ & $0.94 \pm 0.01$ & $0.92 \pm 0.01$ & $0.94 \pm 0.01$ & $0.93 \pm 0.01$ & $0.85 \pm 0.02$ & $0.83 \pm 0.02$ & $0.78 \pm 0.02$ & $0.74 \pm 0.01$ & $0.81 \pm 0.02$ & $0.13 \pm 0.03$ & $0.09 \pm 0.02$ & $0.08 \pm 0.03$ & $0.10 \pm 0.02$ & $0.12 \pm 0.02$ & $0.12 \pm 0.02$ \\ 
  \hline
  12 & Basic & $1.00 \pm 0.00$ & $1.00 \pm 0.00$ & $1.00 \pm 0.00$ & $1.00 \pm 0.00$ & $1.00 \pm 0.00$ & $0.97 \pm 0.01$ & $0.96 \pm 0.01$ & $0.95 \pm 0.01$ & $0.96 \pm 0.01$ & $0.95 \pm 0.01$ & $0.87 \pm 0.01$ & $0.86 \pm 0.01$ & $0.79 \pm 0.01$ & $0.76 \pm 0.01$ & $0.82 \pm 0.01$ & $0.05 \pm 0.01$ & $0.01 \pm 0.00$ & $0.02 \pm 0.00$ & $0.02 \pm 0.01$ & $0.03 \pm 0.01$ & $0.04 \pm 0.01$ \\ 
  12 & Efron & $1.00 \pm 0.00$ & $1.00 \pm 0.00$ & $1.00 \pm 0.00$ & $1.00 \pm 0.00$ & $1.00 \pm 0.00$ & $0.95 \pm 0.01$ & $0.94 \pm 0.01$ & $0.92 \pm 0.01$ & $0.94 \pm 0.01$ & $0.92 \pm 0.01$ & $0.84 \pm 0.02$ & $0.82 \pm 0.01$ & $0.76 \pm 0.01$ & $0.73 \pm 0.01$ & $0.79 \pm 0.01$ & $0.04 \pm 0.01$ & $0.03 \pm 0.01$ & $0.03 \pm 0.01$ & $0.03 \pm 0.01$ & $0.05 \pm 0.01$ & $0.05 \pm 0.01$ \\ 
  12 & BBC & $1.00 \pm 0.00$ & $1.00 \pm 0.00$ & $1.00 \pm 0.00$ & $1.00 \pm 0.00$ & $1.00 \pm 0.00$ & $0.97 \pm 0.01$ & $0.97 \pm 0.01$ & $0.95 \pm 0.01$ & $0.96 \pm 0.01$ & $0.96 \pm 0.01$ & $0.88 \pm 0.02$ & $0.87 \pm 0.01$ & $0.80 \pm 0.01$ & $0.76 \pm 0.01$ & $0.82 \pm 0.01$ & $0.02 \pm 0.01$ & $0.02 \pm 0.01$ & $0.01 \pm 0.00$ & $0.02 \pm 0.00$ & $0.05 \pm 0.01$ & $0.04 \pm 0.01$ \\ 
  \hline
  16 & Basic & $1.00 \pm 0.00$ & $1.00 \pm 0.00$ & $1.00 \pm 0.00$ & $1.00 \pm 0.00$ & $1.00 \pm 0.00$ & $0.98 \pm 0.00$ & $0.97 \pm 0.00$ & $0.96 \pm 0.01$ & $0.96 \pm 0.00$ & $0.96 \pm 0.01$ & $0.88 \pm 0.01$ & $0.87 \pm 0.01$ & $0.80 \pm 0.01$ & $0.76 \pm 0.01$ & $0.82 \pm 0.01$ & $0.02 \pm 0.01$ & $0.01 \pm 0.00$ & $0.01 \pm 0.00$ & $0.01 \pm 0.00$ & $0.01 \pm 0.00$ & $0.02 \pm 0.00$ \\ 
  16 & Efron & $1.00 \pm 0.00$ & $1.00 \pm 0.00$ & $1.00 \pm 0.00$ & $1.00 \pm 0.00$ & $1.00 \pm 0.00$ & $0.97 \pm 0.01$ & $0.96 \pm 0.01$ & $0.94 \pm 0.01$ & $0.95 \pm 0.01$ & $0.94 \pm 0.01$ & $0.86 \pm 0.01$ & $0.84 \pm 0.01$ & $0.77 \pm 0.01$ & $0.74 \pm 0.01$ & $0.80 \pm 0.01$ & $0.02 \pm 0.01$ & $0.01 \pm 0.00$ & $0.01 \pm 0.00$ & $0.01 \pm 0.00$ & $0.02 \pm 0.01$ & $0.03 \pm 0.01$ \\ 
  16 & BBC & $1.00 \pm 0.00$ & $1.00 \pm 0.00$ & $1.00 \pm 0.00$ & $1.00 \pm 0.00$ & $1.00 \pm 0.00$ & $0.99 \pm 0.00$ & $0.99 \pm 0.00$ & $0.99 \pm 0.00$ & $0.99 \pm 0.00$ & $0.99 \pm 0.01$ & $0.92 \pm 0.01$ & $0.91 \pm 0.01$ & $0.83 \pm 0.01$ & $0.79 \pm 0.01$ & $0.85 \pm 0.01$ & $0.00 \pm 0.00$ & $0.00 \pm 0.00$ & $0.00 \pm 0.00$ & $0.00 \pm 0.00$ & $0.01 \pm 0.00$ & $0.01 \pm 0.00$ \\ 
  \hline
  20 & Basic & $1.00 \pm 0.00$ & $1.00 \pm 0.00$ & $1.00 \pm 0.00$ & $1.00 \pm 0.00$ & $1.00 \pm 0.00$ & $0.98 \pm 0.00$ & $0.97 \pm 0.00$ & $0.96 \pm 0.01$ & $0.96 \pm 0.00$ & $0.96 \pm 0.01$ & $0.87 \pm 0.01$ & $0.86 \pm 0.01$ & $0.79 \pm 0.01$ & $0.75 \pm 0.01$ & $0.82 \pm 0.01$ & $0.02 \pm 0.01$ & $0.00 \pm 0.00$ & $0.01 \pm 0.00$ & $0.01 \pm 0.00$ & $0.01 \pm 0.00$ & $0.02 \pm 0.00$ \\ 
  20 & Efron & $1.00 \pm 0.00$ & $1.00 \pm 0.00$ & $1.00 \pm 0.00$ & $1.00 \pm 0.00$ & $1.00 \pm 0.00$ & $0.98 \pm 0.00$ & $0.97 \pm 0.00$ & $0.95 \pm 0.01$ & $0.96 \pm 0.01$ & $0.95 \pm 0.01$ & $0.87 \pm 0.01$ & $0.86 \pm 0.01$ & $0.79 \pm 0.01$ & $0.75 \pm 0.01$ & $0.82 \pm 0.01$ & $0.01 \pm 0.00$ & $0.01 \pm 0.00$ & $0.01 \pm 0.00$ & $0.01 \pm 0.00$ & $0.01 \pm 0.00$ & $0.02 \pm 0.00$ \\ 
  20 & BBC & $1.00 \pm 0.00$ & $1.00 \pm 0.00$ & $1.00 \pm 0.00$ & $1.00 \pm 0.00$ & $1.00 \pm 0.00$ & $1.00 \pm 0.00$ & $1.00 \pm 0.00$ & $1.00 \pm 0.00$ & $1.00 \pm 0.00$ & $1.00 \pm 0.00$ & $0.94 \pm 0.00$ & $0.92 \pm 0.00$ & $0.84 \pm 0.00$ & $0.80 \pm 0.00$ & $0.87 \pm 0.00$ & $0.00 \pm 0.00$ & $0.00 \pm 0.00$ & $0.00 \pm 0.00$ & $0.00 \pm 0.00$ & $0.00 \pm 0.00$ & $0.00 \pm 0.00$ \\ 
\hline
\end{tabular}
}
\caption{Proposed feature importance scores for the overlap dataset. Results are reported for different subspace size $m$, over 20 runs.}
\label{tab:fi_overlap_full}
\end{subtable}

\begin{subtable}{\textwidth}
\centering
\resizebox{\textwidth}{!}{
\begin{tabular}{l|ccccc|ccccc|ccccc|cccccc}
\hline
\textbf{Method} &
$X_1$ & $X_2$ & $X_3$ & $X_4$ & $X_5$ & $X_6$ & $X_7$ & $X_8$ & $X_9$ & $X_{10}$ & $X_{11}$ & $X_{12}$ & $X_{13}$ & $X_{14}$ & $X_{15}$ & $X_{16}$ & $X_{17}$ & $X_{18}$ & $X_{19}$ & $X_{20}$ & $X_{21}$ \\
\hline
URF$_{\text{Gini}}$ &
$0.91\!\pm\!0.08$ & $0.90\!\pm\!0.10$ & $0.85\!\pm\!0.11$ & $0.86\!\pm\!0.11$ & $0.88\!\pm\!0.11$ &
$0.77\!\pm\!0.10$ & $0.75\!\pm\!0.11$ & $0.71\!\pm\!0.11$ & $0.76\!\pm\!0.11$ & $0.74\!\pm\!0.10$ &
$0.63\!\pm\!0.14$ & $0.63\!\pm\!0.12$ & $0.53\!\pm\!0.09$ & $0.54\!\pm\!0.07$ & $0.54\!\pm\!0.09$ &
$0.03\!\pm\!0.02$ & $0.04\!\pm\!0.03$ & $0.02\!\pm\!0.02$ & $0.01\!\pm\!0.02$ & $0.03\!\pm\!0.02$ & $0.03\!\pm\!0.02$ \\

URF$_{\text{perm}}$ &
$0.90\!\pm\!0.30$ & $0.85\!\pm\!0.36$ & $0.85\!\pm\!0.36$ & $0.95\!\pm\!0.22$ & $0.85\!\pm\!0.36$ &
$0.85\!\pm\!0.36$ & $0.85\!\pm\!0.36$ & $0.85\!\pm\!0.36$ & $0.85\!\pm\!0.36$ & $0.85\!\pm\!0.36$ &
$0.85\!\pm\!0.36$ & $0.85\!\pm\!0.36$ & $0.85\!\pm\!0.36$ & $0.85\!\pm\!0.36$ & $0.85\!\pm\!0.36$ &
$0.85\!\pm\!0.36$ & $0.85\!\pm\!0.36$ & $0.85\!\pm\!0.36$ & $0.85\!\pm\!0.36$ & $0.85\!\pm\!0.36$ & $0.85\!\pm\!0.36$ \\

CUBT &
$1.00\!\pm\!0.00$ & $0.99\!\pm\!0.00$ & $0.99\!\pm\!0.00$ & $0.99\!\pm\!0.00$ & $1.00\!\pm\!0.00$ &
$0.94\!\pm\!0.00$ & $0.94\!\pm\!0.00$ & $0.91\!\pm\!0.00$ & $0.92\!\pm\!0.00$ & $0.91\!\pm\!0.00$ &
$0.86\!\pm\!0.00$ & $0.85\!\pm\!0.00$ & $0.80\!\pm\!0.00$ & $0.81\!\pm\!0.00$ & $0.80\!\pm\!0.00$ &
$0.17\!\pm\!0.00$ & $0.15\!\pm\!0.00$ & $0.16\!\pm\!0.00$ & $0.13\!\pm\!0.00$ & $0.10\!\pm\!0.00$ & $0.12\!\pm\!0.00$ \\

TWKM &
$0.77\!\pm\!0.15$ & $0.74\!\pm\!0.20$ & $0.68\!\pm\!0.11$ & $0.94\!\pm\!0.17$ & $0.77\!\pm\!0.15$ &
$0.25\!\pm\!0.05$ & $0.24\!\pm\!0.04$ & $0.17\!\pm\!0.03$ & $0.21\!\pm\!0.03$ & $0.10\!\pm\!0.04$ &
$0.07\!\pm\!0.06$ & $0.08\!\pm\!0.04$ & $0.12\!\pm\!0.03$ & $0.04\!\pm\!0.07$ & $0.06\!\pm\!0.05$ &
$0.02\!\pm\!0.05$ & $0.07\!\pm\!0.22$ & $0.02\!\pm\!0.05$ & $0.01\!\pm\!0.04$ & $0.02\!\pm\!0.05$ & $0.02\!\pm\!0.05$ \\

LS &
$0.24\!\pm\!0.00$ & $0.24\!\pm\!0.00$ & $0.24\!\pm\!0.00$ & $0.24\!\pm\!0.00$ & $0.25\!\pm\!0.00$ &
$0.29\!\pm\!0.00$ & $0.29\!\pm\!0.00$ & $0.30\!\pm\!0.00$ & $0.29\!\pm\!0.00$ & $0.30\!\pm\!0.00$ &
$0.35\!\pm\!0.00$ & $0.35\!\pm\!0.00$ & $0.39\!\pm\!0.00$ & $0.40\!\pm\!0.00$ & $0.36\!\pm\!0.00$ &
$0.98\!\pm\!0.00$ & $0.99\!\pm\!0.00$ & $1.00\!\pm\!0.00$ & $1.00\!\pm\!0.00$ & $0.96\!\pm\!0.00$ & $0.96\!\pm\!0.00$ \\
\hline
\end{tabular}
}
\caption{Feature importance scores of literature methods for the overlap dataset. Results are reported over 50 runs.}
\label{tab:fi_overlap_literature_full}
\end{subtable}
\caption{Comparison of feature importance measures across the proposals and the literature methods considered for the overlap dataset. Results are reported as mean $\pm$ standard deviation.}
\end{sidewaystable}

\begin{sidewaystable}[h]
\centering

\begin{subtable}{\textwidth}
\centering

\resizebox{\textwidth}{!}{%
\begin{tabular}{cl|cccc|cccc|cccccccc}
\hline
$m$ & \textbf{Method} &
$X_1$ & $X_2$ & $X_3$ & $X_4$ & $X_5$ & $X_6$ & $X_7$ & $X_8$ & $X_9$ & $X_{10}$ & $X_{11}$ & $X_{12}$ & $X_{13}$ & $X_{14}$ & $X_{15}$ & $X_{16}$ \\ 
  \hline
9 & Basic & $0.98  \pm  0.00$ & $0.98  \pm  0.00$ & $0.98  \pm  0.00$ & $0.95  \pm  0.01$ & $0.77  \pm  0.01$ & $0.74  \pm  0.01$ & $0.88  \pm  0.01$ & $0.77  \pm  0.01$ & $0.18  \pm  0.02$ & $0.15  \pm  0.02$ & $0.20  \pm  0.02$ & $0.15  \pm  0.01$ & $0.11  \pm  0.01$ & $0.11  \pm  0.02$ & $0.11  \pm  0.01$ & $0.22  \pm  0.02$ \\ 
  9 & Efron & $0.97  \pm  0.01$ & $0.98  \pm  0.00$ & $0.97  \pm  0.00$ & $0.96  \pm  0.01$ & $0.78  \pm  0.01$ & $0.75  \pm  0.01$ & $0.87  \pm  0.01$ & $0.78  \pm  0.01$ & $0.18  \pm  0.02$ & $0.16  \pm  0.01$ & $0.20  \pm  0.02$ & $0.16  \pm  0.02$ & $0.13  \pm  0.02$ & $0.12  \pm  0.02$ & $0.12  \pm  0.02$ & $0.22  \pm  0.02$ \\ 
  9 & BBC & $0.97  \pm  0.01$ & $0.98  \pm  0.00$ & $0.97  \pm  0.00$ & $0.96  \pm  0.01$ & $0.79  \pm  0.01$ & $0.75  \pm  0.01$ & $0.87  \pm  0.01$ & $0.77  \pm  0.01$ & $0.19  \pm  0.02$ & $0.15  \pm  0.02$ & $0.21  \pm  0.03$ & $0.16  \pm  0.02$ & $0.14  \pm  0.01$ & $0.13  \pm  0.01$ & $0.13  \pm  0.02$ & $0.23  \pm  0.02$ \\ 
  \hline
  11 & Basic & $0.98  \pm  0.00$ & $0.99  \pm  0.00$ & $0.98  \pm  0.00$ & $0.95  \pm  0.01$ & $0.74  \pm  0.01$ & $0.70  \pm  0.01$ & $0.87  \pm  0.01$ & $0.74  \pm  0.01$ & $0.13  \pm  0.01$ & $0.08  \pm  0.01$ & $0.15  \pm  0.01$ & $0.10  \pm  0.01$ & $0.08  \pm  0.01$ & $0.06  \pm  0.01$ & $0.06  \pm  0.01$ & $0.15  \pm  0.01$ \\ 
  11 & Efron & $0.97  \pm  0.00$ & $0.98  \pm  0.00$ & $0.97  \pm  0.00$ & $0.96  \pm  0.01$ & $0.75  \pm  0.01$ & $0.72  \pm  0.01$ & $0.86  \pm  0.01$ & $0.75  \pm  0.01$ & $0.11  \pm  0.01$ & $0.09  \pm  0.01$ & $0.13  \pm  0.02$ & $0.09  \pm  0.01$ & $0.08  \pm  0.01$ & $0.07  \pm  0.01$ & $0.07  \pm  0.01$ & $0.15  \pm  0.02$ \\ 
  11 & BBC & $0.97  \pm  0.00$ & $0.98  \pm  0.00$ & $0.97  \pm  0.00$ & $0.96  \pm  0.01$ & $0.76  \pm  0.01$ & $0.72  \pm  0.01$ & $0.85  \pm  0.01$ & $0.75  \pm  0.01$ & $0.12  \pm  0.01$ & $0.09  \pm  0.01$ & $0.13  \pm  0.02$ & $0.09  \pm  0.01$ & $0.09  \pm  0.01$ & $0.07  \pm  0.01$ & $0.07  \pm  0.01$ & $0.16  \pm  0.01$ \\ 
  \hline
  13 & Basic & $0.99  \pm  0.00$ & $0.99  \pm  0.00$ & $0.98  \pm  0.00$ & $0.95  \pm  0.01$ & $0.73  \pm  0.01$ & $0.69  \pm  0.01$ & $0.87  \pm  0.01$ & $0.73  \pm  0.01$ & $0.10  \pm  0.01$ & $0.05  \pm  0.00$ & $0.11  \pm  0.01$ & $0.07  \pm  0.01$ & $0.06  \pm  0.01$ & $0.03  \pm  0.01$ & $0.04  \pm  0.00$ & $0.12  \pm  0.01$ \\ 
  13 & Efron & $0.98  \pm  0.00$ & $0.98  \pm  0.00$ & $0.98  \pm  0.00$ & $0.95  \pm  0.01$ & $0.73  \pm  0.01$ & $0.69  \pm  0.01$ & $0.85  \pm  0.01$ & $0.73  \pm  0.01$ & $0.08  \pm  0.01$ & $0.05  \pm  0.01$ & $0.09  \pm  0.01$ & $0.06  \pm  0.01$ & $0.06  \pm  0.01$ & $0.04  \pm  0.01$ & $0.04  \pm  0.00$ & $0.12  \pm  0.01$ \\ 
  13 & BBC & $0.97  \pm  0.01$ & $0.98  \pm  0.00$ & $0.97  \pm  0.00$ & $0.95  \pm  0.00$ & $0.73  \pm  0.01$ & $0.69  \pm  0.01$ & $0.84  \pm  0.01$ & $0.72  \pm  0.01$ & $0.09  \pm  0.01$ & $0.06  \pm  0.01$ & $0.09  \pm  0.01$ & $0.06  \pm  0.01$ & $0.06  \pm  0.01$ & $0.04  \pm  0.01$ & $0.04  \pm  0.01$ & $0.12  \pm  0.01$ \\ 
  \hline
  15 & Basic & $0.99  \pm  0.00$ & $0.99  \pm  0.00$ & $0.99  \pm  0.00$ & $0.96  \pm  0.00$ & $0.74  \pm  0.01$ & $0.69  \pm  0.01$ & $0.88  \pm  0.01$ & $0.73  \pm  0.01$ & $0.08  \pm  0.01$ & $0.03  \pm  0.01$ & $0.08  \pm  0.01$ & $0.05  \pm  0.01$ & $0.06  \pm  0.01$ & $0.02  \pm  0.00$ & $0.03  \pm  0.00$ & $0.09  \pm  0.01$ \\ 
  15 & Efron & $0.98  \pm  0.00$ & $0.98  \pm  0.00$ & $0.98  \pm  0.00$ & $0.96  \pm  0.01$ & $0.72  \pm  0.01$ & $0.68  \pm  0.01$ & $0.85  \pm  0.01$ & $0.72  \pm  0.01$ & $0.07  \pm  0.01$ & $0.04  \pm  0.01$ & $0.07  \pm  0.01$ & $0.05  \pm  0.01$ & $0.05  \pm  0.00$ & $0.03  \pm  0.00$ & $0.03  \pm  0.00$ & $0.09  \pm  0.01$ \\ 
  15 & BBC & $0.97  \pm  0.00$ & $0.98  \pm  0.00$ & $0.97  \pm  0.00$ & $0.95  \pm  0.01$ & $0.72  \pm  0.01$ & $0.68  \pm  0.01$ & $0.83  \pm  0.00$ & $0.71  \pm  0.01$ & $0.07  \pm  0.01$ & $0.04  \pm  0.01$ & $0.07  \pm  0.01$ & $0.05  \pm  0.01$ & $0.05  \pm  0.01$ & $0.03  \pm  0.00$ & $0.03  \pm  0.00$ & $0.09  \pm  0.01$ \\ 
   \hline
\end{tabular}
}
\caption{Proposed feature importance scores for the proportion dataset. Results are reported for different subspace size $m$, over 20 runs.}
\label{tab:fi_proportion_full}
\end{subtable}

\begin{subtable}{\textwidth}
\centering
\resizebox{\textwidth}{!}{
\begin{tabular}{l|cccc|cccc|cccccccc}
\hline
\textbf{Method} &
$X_1$ & $X_2$ & $X_3$ & $X_4$ & $X_5$ & $X_6$ & $X_7$ & $X_8$ & $X_9$ & $X_{10}$ & $X_{11}$ & $X_{12}$ & $X_{13}$ & $X_{14}$ & $X_{15}$ & $X_{16}$ \\
\hline
URF$_{\text{Gini}}$ &
$0.90\!\pm\!0.08$ & $0.88\!\pm\!0.10$ & $0.85\!\pm\!0.10$ & $0.91\!\pm\!0.10$ &
$0.63\!\pm\!0.13$ & $0.59\!\pm\!0.09$ & $0.76\!\pm\!0.11$ & $0.63\!\pm\!0.10$ &
$0.04\!\pm\!0.03$ & $0.03\!\pm\!0.02$ & $0.05\!\pm\!0.03$ & $0.04\!\pm\!0.03$ &
$0.04\!\pm\!0.03$ & $0.03\!\pm\!0.02$ & $0.03\!\pm\!0.03$ & $0.07\!\pm\!0.03$ \\

URF$_{\text{perm}}$ &
$0.56\!\pm\!0.29$ & $0.60\!\pm\!0.28$ & $0.50\!\pm\!0.34$ & $0.72\!\pm\!0.32$ &
$0.11\!\pm\!0.15$ & $0.05\!\pm\!0.06$ & $0.30\!\pm\!0.25$ & $0.15\!\pm\!0.17$ &
$0.00\!\pm\!0.00$ & $0.00\!\pm\!0.00$ & $0.00\!\pm\!0.00$ & $0.00\!\pm\!0.00$ &
$0.00\!\pm\!0.00$ & $0.00\!\pm\!0.00$ & $0.00\!\pm\!0.00$ & $0.00\!\pm\!0.00$ \\

CUBT &
$1.00\!\pm\!0.00$ & $0.99\!\pm\!0.00$ & $0.98\!\pm\!0.00$ & $1.00\!\pm\!0.00$ &
$0.86\!\pm\!0.00$ & $0.88\!\pm\!0.00$ & $0.88\!\pm\!0.00$ & $0.84\!\pm\!0.00$ &
$0.15\!\pm\!0.00$ & $0.16\!\pm\!0.00$ & $0.21\!\pm\!0.00$ & $0.17\!\pm\!0.00$ &
$0.24\!\pm\!0.00$ & $0.12\!\pm\!0.00$ & $0.22\!\pm\!0.00$ & $0.31\!\pm\!0.00$ \\

TWKM &
$0.82\!\pm\!0.27$ & $0.79\!\pm\!0.21$ & $0.68\!\pm\!0.22$ & $0.80\!\pm\!0.28$ &
$0.35\!\pm\!0.18$ & $0.20\!\pm\!0.11$ & $0.34\!\pm\!0.17$ & $0.15\!\pm\!0.09$ &
$0.04\!\pm\!0.07$ & $0.06\!\pm\!0.09$ & $0.14\!\pm\!0.30$ & $0.05\!\pm\!0.09$ &
$0.05\!\pm\!0.08$ & $0.04\!\pm\!0.08$ & $0.05\!\pm\!0.08$ & $0.10\!\pm\!0.22$ \\

LS &
$0.51\!\pm\!0.00$ & $0.50\!\pm\!0.00$ & $0.51\!\pm\!0.00$ & $0.50\!\pm\!0.00$ &
$0.59\!\pm\!0.00$ & $0.61\!\pm\!0.00$ & $0.55\!\pm\!0.00$ & $0.59\!\pm\!0.00$ &
$0.98\!\pm\!0.00$ & $0.99\!\pm\!0.00$ & $0.98\!\pm\!0.00$ & $0.99\!\pm\!0.00$ &
$0.99\!\pm\!0.00$ & $1.00\!\pm\!0.00$ & $1.00\!\pm\!0.00$ & $0.95\!\pm\!0.00$ \\
\hline
\end{tabular}
}
\caption{Feature importance scores of literature methods for the proportion dataset. Results are reported over 50 runs.}
\label{tab:fi_proportion_literature_full}
\end{subtable}
\caption{Comparison of feature importance measures across the proposals and the literature methods considered for the proportion dataset. Results are reported as mean $\pm$ standard deviation.}
\end{sidewaystable}

\begin{sidewaystable}[h]
\centering
\begin{subtable}{\textwidth}
\centering
\resizebox{\textwidth}{!}{%
\begin{tabular}{cl|cccc|cccc|cccc|cccccc}
\hline
$m$ & \textbf{Method} &
$X_1$ & $X_2$ & $X_3$ & $X_4$ & $X_5$ & $X_6$ & $X_7$ & $X_8$ & $X_9$ & $X_{10}$ & $X_{11}$ & $X_{12}$ & $X_{13}$ & $X_{14}$ & $X_{15}$ & $X_{16}$ & $X_{17}$ & $X_{18}$ \\ 
\hline
9 & Basic & $1.00  \pm  0.00$ & $0.99  \pm  0.00$ & $0.97  \pm  0.00$ & $0.98  \pm  0.00$ & $0.93  \pm  0.01$ & $0.96  \pm  0.01$ & $0.95  \pm  0.01$ & $0.95  \pm  0.01$ & $0.89  \pm  0.01$ & $0.90  \pm  0.01$ & $0.86  \pm  0.01$ & $0.86  \pm  0.01$ & $0.08  \pm  0.01$ & $0.08  \pm  0.01$ & $0.10  \pm  0.01$ & $0.09  \pm  0.02$ & $0.12  \pm  0.02$ & $0.06  \pm  0.02$ \\ 
  9 & Efron & $0.99  \pm  0.00$ & $0.99  \pm  0.00$ & $0.96  \pm  0.01$ & $0.98  \pm  0.01$ & $0.92  \pm  0.01$ & $0.94  \pm  0.01$ & $0.93  \pm  0.01$ & $0.93  \pm  0.01$ & $0.87  \pm  0.01$ & $0.88  \pm  0.01$ & $0.84  \pm  0.01$ & $0.84  \pm  0.01$ & $0.08  \pm  0.02$ & $0.09  \pm  0.01$ & $0.11  \pm  0.02$ & $0.10  \pm  0.02$ & $0.11  \pm  0.02$ & $0.09  \pm  0.02$ \\ 
  9 & BBC & $1.00  \pm  0.00$ & $0.99  \pm  0.00$ & $0.97  \pm  0.01$ & $0.98  \pm  0.00$ & $0.93  \pm  0.01$ & $0.96  \pm  0.01$ & $0.94  \pm  0.01$ & $0.95  \pm  0.01$ & $0.88  \pm  0.01$ & $0.90  \pm  0.01$ & $0.86  \pm  0.02$ & $0.86  \pm  0.01$ & $0.07  \pm  0.01$ & $0.08  \pm  0.01$ & $0.09  \pm  0.01$ & $0.09  \pm  0.02$ & $0.10  \pm  0.02$ & $0.07  \pm  0.02$ \\ 
  \hline
  11 & Basic & $1.00  \pm  0.00$ & $1.00  \pm  0.00$ & $0.98  \pm  0.00$ & $0.99  \pm  0.00$ & $0.95  \pm  0.01$ & $0.97  \pm  0.01$ & $0.96  \pm  0.01$ & $0.96  \pm  0.01$ & $0.91  \pm  0.01$ & $0.92  \pm  0.01$ & $0.88  \pm  0.01$ & $0.87  \pm  0.01$ & $0.04  \pm  0.01$ & $0.05  \pm  0.01$ & $0.05  \pm  0.01$ & $0.04  \pm  0.01$ & $0.06  \pm  0.02$ & $0.02  \pm  0.01$ \\ 
  11 & Efron & $1.00  \pm  0.00$ & $0.99  \pm  0.00$ & $0.97  \pm  0.00$ & $0.98  \pm  0.00$ & $0.92  \pm  0.01$ & $0.95  \pm  0.01$ & $0.93  \pm  0.01$ & $0.93  \pm  0.01$ & $0.88  \pm  0.01$ & $0.88  \pm  0.01$ & $0.84  \pm  0.01$ & $0.83  \pm  0.01$ & $0.05  \pm  0.01$ & $0.06  \pm  0.01$ & $0.07  \pm  0.01$ & $0.05  \pm  0.01$ & $0.06  \pm  0.01$ & $0.04  \pm  0.01$ \\ 
  11 & BBC & $1.00  \pm  0.00$ & $0.99  \pm  0.00$ & $0.97  \pm  0.01$ & $0.98  \pm  0.00$ & $0.93  \pm  0.01$ & $0.96  \pm  0.01$ & $0.95  \pm  0.01$ & $0.95  \pm  0.01$ & $0.89  \pm  0.01$ & $0.90  \pm  0.01$ & $0.86  \pm  0.01$ & $0.85  \pm  0.01$ & $0.04  \pm  0.01$ & $0.06  \pm  0.01$ & $0.05  \pm  0.01$ & $0.05  \pm  0.01$ & $0.05  \pm  0.01$ & $0.04  \pm  0.01$ \\ 
  \hline
  13 & Basic & $1.00  \pm  0.00$ & $1.00  \pm  0.00$ & $0.98  \pm  0.00$ & $0.99  \pm  0.00$ & $0.96  \pm  0.01$ & $0.98  \pm  0.00$ & $0.96  \pm  0.01$ & $0.97  \pm  0.00$ & $0.93  \pm  0.01$ & $0.94  \pm  0.01$ & $0.90  \pm  0.01$ & $0.89  \pm  0.01$ & $0.02  \pm  0.01$ & $0.04  \pm  0.00$ & $0.03  \pm  0.00$ & $0.02  \pm  0.00$ & $0.03  \pm  0.01$ & $0.01  \pm  0.00$ \\ 
  13 & Efron & $1.00  \pm  0.00$ & $1.00  \pm  0.00$ & $0.97  \pm  0.00$ & $0.98  \pm  0.00$ & $0.93  \pm  0.01$ & $0.96  \pm  0.01$ & $0.94  \pm  0.01$ & $0.94  \pm  0.01$ & $0.88  \pm  0.01$ & $0.90  \pm  0.01$ & $0.86  \pm  0.01$ & $0.84  \pm  0.01$ & $0.03  \pm  0.01$ & $0.05  \pm  0.00$ & $0.04  \pm  0.01$ & $0.03  \pm  0.01$ & $0.04  \pm  0.01$ & $0.02  \pm  0.01$ \\ 
  13 & BBC & $1.00  \pm  0.00$ & $1.00  \pm  0.00$ & $0.98  \pm  0.00$ & $0.99  \pm  0.00$ & $0.95  \pm  0.01$ & $0.97  \pm  0.01$ & $0.96  \pm  0.01$ & $0.96  \pm  0.01$ & $0.91  \pm  0.01$ & $0.92  \pm  0.01$ & $0.88  \pm  0.01$ & $0.86  \pm  0.01$ & $0.02  \pm  0.00$ & $0.04  \pm  0.01$ & $0.03  \pm  0.01$ & $0.02  \pm  0.00$ & $0.02  \pm  0.01$ & $0.02  \pm  0.00$ \\ 
  \hline
  15 & Basic & $1.00  \pm  0.00$ & $1.00  \pm  0.00$ & $0.99  \pm  0.00$ & $1.00  \pm  0.00$ & $0.97  \pm  0.00$ & $0.98  \pm  0.00$ & $0.97  \pm  0.01$ & $0.98  \pm  0.00$ & $0.95  \pm  0.01$ & $0.96  \pm  0.01$ & $0.92  \pm  0.01$ & $0.91  \pm  0.01$ & $0.01  \pm  0.00$ & $0.04  \pm  0.00$ & $0.01  \pm  0.00$ & $0.01  \pm  0.00$ & $0.01  \pm  0.00$ & $0.01  \pm  0.00$ \\ 
  15 & Efron & $1.00  \pm  0.00$ & $1.00  \pm  0.00$ & $0.98  \pm  0.00$ & $0.99  \pm  0.00$ & $0.94  \pm  0.01$ & $0.96  \pm  0.01$ & $0.95  \pm  0.01$ & $0.95  \pm  0.01$ & $0.90  \pm  0.01$ & $0.91  \pm  0.01$ & $0.86  \pm  0.01$ & $0.85  \pm  0.01$ & $0.02  \pm  0.00$ & $0.04  \pm  0.01$ & $0.02  \pm  0.01$ & $0.02  \pm  0.00$ & $0.02  \pm  0.01$ & $0.02  \pm  0.00$ \\ 
  15 & BBC & $1.00  \pm  0.00$ & $1.00  \pm  0.00$ & $0.99  \pm  0.00$ & $0.99  \pm  0.00$ & $0.96  \pm  0.01$ & $0.98  \pm  0.01$ & $0.97  \pm  0.01$ & $0.97  \pm  0.01$ & $0.92  \pm  0.01$ & $0.94  \pm  0.01$ & $0.90  \pm  0.01$ & $0.88  \pm  0.01$ & $0.01  \pm  0.00$ & $0.04  \pm  0.00$ & $0.01  \pm  0.00$ & $0.01  \pm  0.00$ & $0.01  \pm  0.00$ & $0.01  \pm  0.00$ \\ 
   \hline
\end{tabular}
}
\caption{Proposed feature importance scores for the correlation dataset. Results are reported for different subspace size $m$, over 20 runs.}
\label{tab:fi_correlation_full}
\end{subtable}

\begin{subtable}{\textwidth}
\centering
\resizebox{\textwidth}{!}{
\begin{tabular}{l|cccc|cccc|cccc|cccccc}
\hline
\textbf{Method} &
$X_1$ & $X_2$ & $X_3$ & $X_4$ & $X_5$ & $X_6$ & $X_7$ & $X_8$ & $X_9$ & $X_{10}$ & $X_{11}$ & $X_{12}$ & $X_{13}$ & $X_{14}$ & $X_{15}$ & $X_{16}$ & $X_{17}$ & $X_{18}$ \\
\hline
URF$_{\text{Gini}}$ &
$0.90\!\pm\!0.09$ & $0.89\!\pm\!0.09$ & $0.82\!\pm\!0.10$ & $0.83\!\pm\!0.14$ &
$0.80\!\pm\!0.10$ & $0.82\!\pm\!0.11$ & $0.79\!\pm\!0.12$ & $0.78\!\pm\!0.13$ &
$0.81\!\pm\!0.13$ & $0.78\!\pm\!0.12$ & $0.69\!\pm\!0.11$ & $0.63\!\pm\!0.12$ &
$0.01\!\pm\!0.02$ & $0.03\!\pm\!0.01$ & $0.03\!\pm\!0.02$ & $0.02\!\pm\!0.01$ &
$0.03\!\pm\!0.02$ & $0.02\!\pm\!0.02$ \\

URF$_{\text{perm}}$ &
$0.32\!\pm\!0.42$ & $0.31\!\pm\!0.38$ & $0.13\!\pm\!0.32$ & $0.16\!\pm\!0.31$ &
$0.28\!\pm\!0.43$ & $0.20\!\pm\!0.36$ & $0.16\!\pm\!0.36$ & $0.22\!\pm\!0.37$ &
$0.13\!\pm\!0.31$ & $0.18\!\pm\!0.32$ & $0.12\!\pm\!0.30$ & $0.06\!\pm\!0.22$ &
$0.05\!\pm\!0.22$ & $0.05\!\pm\!0.22$ & $0.05\!\pm\!0.22$ & $0.05\!\pm\!0.22$ &
$0.05\!\pm\!0.22$ & $0.05\!\pm\!0.22$ \\

CUBT &
$0.97\!\pm\!0.00$ & $1.00\!\pm\!0.00$ & $0.99\!\pm\!0.00$ & $0.97\!\pm\!0.00$ &
$0.92\!\pm\!0.00$ & $0.95\!\pm\!0.00$ & $0.92\!\pm\!0.00$ & $0.93\!\pm\!0.00$ &
$0.91\!\pm\!0.00$ & $0.90\!\pm\!0.00$ & $0.90\!\pm\!0.00$ & $0.87\!\pm\!0.00$ &
$0.21\!\pm\!0.00$ & $0.21\!\pm\!0.00$ & $0.17\!\pm\!0.00$ & $0.22\!\pm\!0.00$ &
$0.21\!\pm\!0.00$ & $0.13\!\pm\!0.00$ \\

TWKM &
$0.81\!\pm\!0.13$ & $0.86\!\pm\!0.25$ & $0.57\!\pm\!0.19$ & $0.76\!\pm\!0.21$ &
$0.19\!\pm\!0.15$ & $0.28\!\pm\!0.17$ & $0.40\!\pm\!0.18$ & $0.33\!\pm\!0.11$ &
$0.10\!\pm\!0.02$ & $0.14\!\pm\!0.02$ & $0.09\!\pm\!0.02$ & $0.23\!\pm\!0.07$ &
$0.02\!\pm\!0.04$ & $0.02\!\pm\!0.04$ & $0.02\!\pm\!0.04$ & $0.01\!\pm\!0.03$ &
$0.02\!\pm\!0.04$ & $0.02\!\pm\!0.05$ \\

LS &
$0.28\!\pm\!0.00$ & $0.29\!\pm\!0.00$ & $0.31\!\pm\!0.00$ & $0.30\!\pm\!0.00$ &
$0.32\!\pm\!0.00$ & $0.30\!\pm\!0.00$ & $0.32\!\pm\!0.00$ & $0.32\!\pm\!0.00$ &
$0.34\!\pm\!0.00$ & $0.33\!\pm\!0.00$ & $0.36\!\pm\!0.00$ & $0.37\!\pm\!0.00$ &
$0.99\!\pm\!0.00$ & $1.00\!\pm\!0.00$ & $0.99\!\pm\!0.00$ & $0.98\!\pm\!0.00$ &
$0.99\!\pm\!0.00$ & $0.95\!\pm\!0.00$ \\
\hline
\end{tabular}
}
\caption{Feature importance scores of literature methods for the correlation dataset. Results are reported over 50 runs.}
\label{tab:fi_correlation_literature_full}
\end{subtable}
\caption{Comparison of feature importance measures across the proposals and the literature methods considered for the correlation dataset. Results are reported as mean $\pm$ standard deviation.}
\end{sidewaystable}

\end{document}